\documentclass[10pt,twocolumn,letterpaper]{article}

\usepackage{cvpr}              
\usepackage{amsmath, amssymb, algorithm, algpseudocode}
\usepackage{pifont}

\newcommand{\accident}{\normalsize\ding{83}}
\usepackage{graphicx}
\usepackage{float} 
\usepackage{pgfplots}
\usepackage{multirow}
\usepackage{balance}

\definecolor{cvprblue}{rgb}{0.21,0.49,0.74}
\usepackage[pagebackref,breaklinks,colorlinks,allcolors=cvprblue]{hyperref}

\def\paperID{3319} 
\def\confName{CVPR}
\def\confYear{2025}

\title{GEM: A Generalizable Ego-Vision Multimodal World Model for Fine-Grained Ego-Motion, Object Dynamics, and Scene Composition Control}

\author{ Mariam Hassan$^\star$$^1$, Sebastian Stapf$^\star$$^2$, Ahmad Rahimi$^\star$$^1$, Pedro M B Rezende$^\star$$^2$, Yasaman Haghighi$^\Diamond$$^1$, \\
David Brüggemann$^\Diamond$$^3$, Isinsu Katircioglu$^\Diamond$$^3$, Lin Zhang$^\Diamond$$^3$, Xiaoran Chen$^\Diamond$$^3$, Suman Saha$^\Diamond$$^3$, \\
Marco Cannici$^\Diamond$$^4$, Elie Aljalbout$^\Diamond$$^4$, Botao Ye$^\Diamond$$^5$, Xi Wang$^\Diamond$$^5$, Aram Davtyan$^2$,  \\
Mathieu Salzmann$^{1,3}$, Davide Scaramuzza$^4$, Marc Pollefeys$^5$, Paolo Favaro$^2$, Alexandre Alahi$^1$\\ \\
$^1$ École Polytechnique Fédérale de Lausanne (EPFL), $^2$ University of Bern, \\ 
$^3$ Swiss Data Science Center, $^4$ University of Zurich, $^5$ ETH Zurich \\
}

\begin{document}
\twocolumn[{%
\renewcommand\twocolumn[1][]{#1}%
\maketitle

\begin{center}
    \captionsetup{type=figure}
    \includegraphics[width=1.0\textwidth]{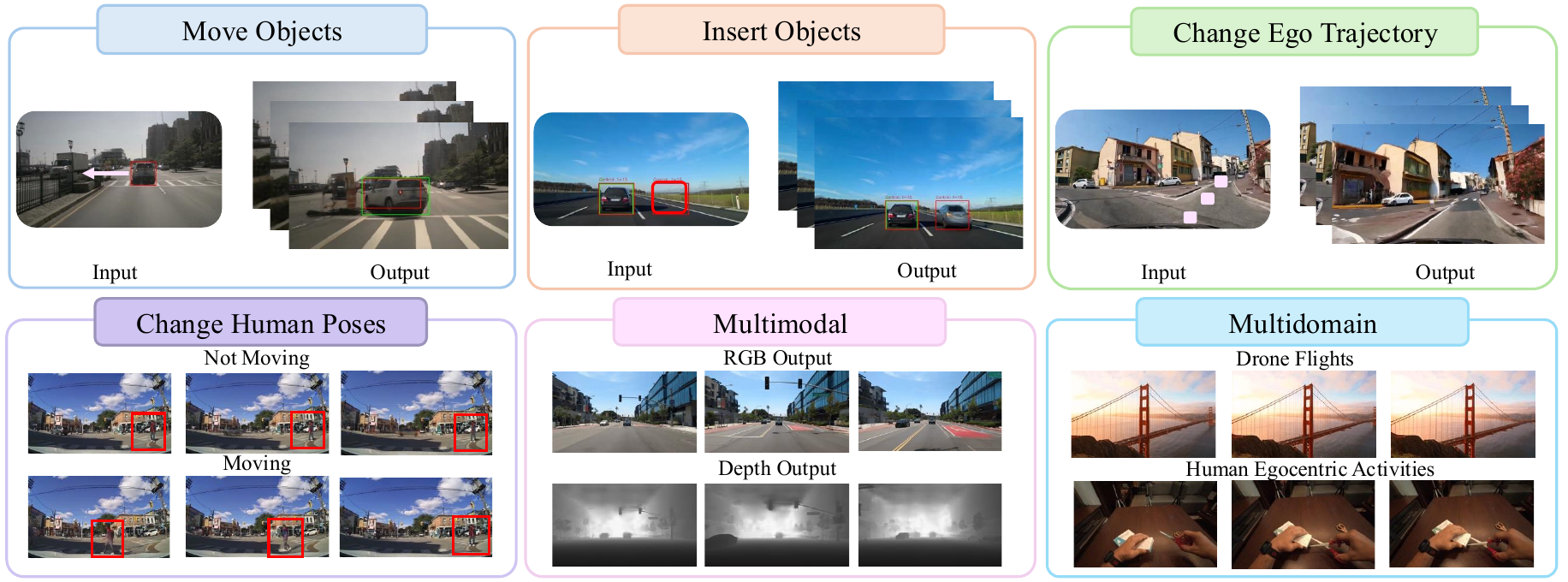}
    \caption{Overview of the capabilities of our proposed world model. GEM enables a range of features, including object manipulation (move, and insert objects), dynamic ego-trajectory adjustments, human poses changes and adaptability to multimodal outputs (i.e., images and depth maps) and multiple domains (i.e., drones and human egocentric activities). All images are generated by GEM.}

    \label{fig:pull_figure}
\end{center}
}]

\begingroup
\renewcommand\thefootnote{}
\footnotetext{$^\star$Main Contributors, $^\Diamond$  Data Contributors }
\endgroup
\begin{abstract}
We present \textbf{GEM}, a \textbf{G}eneralizable \textbf{E}go-vision \textbf{M}ultimodal world model that predicts future frames using a reference frame, sparse features, human poses, and ego-trajectories. Hence, our model has precise control over object dynamics, ego-agent motion and human poses. GEM generates paired RGB and depth outputs for richer spatial understanding. We introduce autoregressive noise schedules to enable stable long-horizon generations. 
Our dataset is comprised of 4000+ hours of multimodal data across domains like autonomous driving, egocentric human activities, and drone flights. 
Pseudo-labels are used to get depth maps, ego-trajectories, and human poses. 
We use a comprehensive evaluation framework, including a new Control of Object Manipulation (COM) metric, to assess controllability.
Experiments show GEM excels at generating diverse, controllable scenarios and temporal consistency over long generations. Code, models, and datasets are fully open-sourced\footnote{\url{https://vita-epfl.github.io/GEM.github.io/}}.

\end{abstract}    

\section{Introduction} 
\label{sec:intro}
Different ego-vision tasks, such as autonomous driving, egocentric human activities, and drone navigation, share a common set of challenges centered around understanding and interacting with the environment from a first-person perspective. Whether it is a car moving, a drone flying, or a human preparing a meal, ego-agents are inherently highly dynamic and interactive within their environment. Consequently, planning for ego-vision tasks requires understanding the dynamics and interactions occurring in respective environments, as well as understanding the effects that ego-agent's actions have on their surroundings.

World models predict plausible futures given past observations and control signals~\cite{hu2023generalpurposerobotsfoundationmodels, LeCun}. By replicating the distribution of appearances and dynamics in observed visual data, they capture patterns and principles that drive interactions. This imaginative capability makes them excellent tools for decision-making in ego-vision tasks~\cite{hu2023generalpurposerobotsfoundationmodels}.
Existing egocentric world models~\cite{kim2021drivegan,drivedreamer,drivedreamer2,wovogen, drivewm, jia2023adriver, vista} perform well with different controls but mainly focus on a single ego-vision task, such as autonomous driving, with domain-dependent control technique. The control in such models is primarily egocentric, \ie, only capturing the motion and actions of the ego-agent. This limits the diversity of the generated scenes, making it hard to model complex interactions such as changing agents' locations in the scene. Overcoming such limitations comes with key challenges in scaling datasets, generalizing controls, and developing tailored evaluation frameworks for controllability.

To address the gaps highlighted above, we propose GEM, a \textbf{G}eneralizable \textbf{E}go-vision \textbf{M}ultimodal world model with high-fidelity controls. As summarized in~\cref{fig:pull_figure},
GEM is a multimodal and multidomain model designed to adapt to different ego-vision tasks while enabling fine-grained control over the scene in an unsupervised manner. GEM's controlling technique is threefold: 
1) ego-motion control through ego trajectories,  2) scene composition control by inpainting the future content and the dynamics from a sparse set of visual tokens, and 3) a more fine-grained control over human motions through human poses. For scene composition control, we use sparse visual tokens extracted by the DINOv2~\cite{dinov2} encoder augmented with unique object identification codes to enable precise control over the motion and appearance of all objects in the scene. We support the insertion of entirely new objects, and allow for highly flexible and controllable future prediction. 
Alongside the frames, GEM is capable of generating depth providing rich spatial context. To achieve that, we pseudo-label our data with depth maps, ego-trajectories and human poses.

GEM is trained on a large corpus of open-source datasets, with contributions in both methods and datasets. The methods  are summarized as follows:

\begin{itemize}

\item We present GEM, a generalizable world model that predicts future frames given a reference, sparse DINOv2 features, human pose, and ego-trajectories. It enables control over ego motion, object dynamics, and human poses. 
\item We introduce autoregressive noise schedules to our framework, enabling stable long-horizon generations.
\item We train on autonomous driving domain and explore multimodal and multidomain generation by (1) integrating depth as an extra generation modality; and (2) fine-tuning our model on different ego-vision domains, \ie, human ego activities, and drone navigations.
\item We present a comprehensive evaluation of GEM's controllability and introduce a metric, Control of Object Manipulation (COM) to evaluate control of object motion.


\end{itemize}

To address limitations in scale and diversity of existing open-source datasets, we propose the following dataset-related contributions:
\begin{itemize}
    \item We utilize a large-scale open-source corpus with over 3200 hours of driving videos, 1000 hours of egocentric human activity datasets, and 27.4 hours of self-collected drone footage from YouTube. The driving datasets are further curated for diverse interactions and dynamics.
    \item Given scarcity of labels, we implement pseudo-labeling approaches to generate depth maps, ego-trajectories, and human pose annotations. We show the effectiveness of our control strategy given pseudo-labels.  
\end{itemize}
Our work is fully open-source, sharing the curated datasets, codebase, and models\footnote{\url{https://vita-epfl.github.io/GEM.github.io/}}.

\section{Related Work} 
\label{sec:related_works}
We briefly review the previous works on controllable video generation models and world models including autonomous driving and egocentric human activity world models. \\

\noindent \textbf{Controllable Video Generation.}
Recent advancements in video generation models have enabled realistic, high-quality video rendering. Several pioneering models leverage Large Language Models (LLMs) for text-to-video generation~\cite{lu2024fitflexiblevisiontransformer, yan2021videogptvideogenerationusing}. Since the success of diffusion models~\cite{rombach2022highresolutionimagesynthesislatent, dhariwal2021diffusionmodelsbeatgans}, diffusion-based video generation has become prominent. Methods can be categorized as: text-to-video~\cite{guo2024animatediffanimatepersonalizedtexttoimage, ho2022videodiffusionmodels, singer2022makeavideotexttovideogenerationtextvideo, voleti2022mcvdmaskedconditionalvideo, wang2024videocomposer, chen2024videocrafter2, ho2022imagen} or image-to-video, ~\cite{svd,zhang2023i2vgenxlhighqualityimagetovideosynthesis, chen2023videocrafter1opendiffusionmodels}. Diffusion models adapt to various control inputs like text, edge maps, and depth maps~\cite{controlnet}; they also offer superior realism~\cite{svd}. However, generic video generation models are not trained to encode the intricate dynamics of egocentric environments~\cite{genad}, and many do not offer detailed motion controls over the generations. \\

\noindent\textbf{World Models.} World models are large-scale generative models that infer dynamics and predict plausible futures based on past observations~\cite{ hu2023generalpurposerobotsfoundationmodels,LeCun, yang2023learning}. 
They are valuable in many tasks such as real-world simulations~\cite{yang2023learning, zhu2024soraworldsimulatorcomprehensive}, reinforcement learning ~\cite{Hafner2020Dream, hao2023reasoning, 8967559, 10.5555/3666122.3667848,2024limt}, model-predictive control~\cite{hansen2022temporal,hansen2024hierarchical}, and representation learning ~\cite{he2024largescaleactionlessvideopretraining, mendonca2023structuredworldmodelshuman}.\\

\noindent\textbf{Autonomous Driving World Models. }
World models for autonomous driving represent the world using sensor observations, such as lidar-generated point clouds~\cite{zhang2024copilot4dlearningunsupervisedworld,bogdoll2024muvomultimodalworldmodel,zheng2023occworld,yang2024visual}, with limited datasets often constraining their scale, or images~\cite{gaia,kim2021drivegan, wovogen, drivewm, drivedreamer, drivedreamer2, genad, vista}. Recent visual world models use LLMs as backbones~\cite{drivedreamer2,jia2023adriver, xiang2024pandora}, but these models rely heavily on LLMs’ spatial reasoning, which remains limited~\cite{hu2023language, zhang2024can, plaat2024reasoning}. This makes them better suited for high-level scene control, like weather or lighting adjustments, rather than precise motion control~\cite{vista}. 
Diffusion-based models, in contrast, use low-level controls like ego-trajectories and maps~\cite{vista, wovogen, drivedreamer, genad, drivewm, drivedreamer2}, but focus  primarily on ego-centric control, limiting their ability to generate complex scenarios such as controlling over any other motions in the scene. Additionally, efforts to improve multimodal world models for spatial metric understanding~\cite{bogdoll2024muvomultimodalworldmodel} rely on limited simulation-based point cloud datasets, which are difficult to generalize to real-world data. \\

\noindent\textbf{Egocentric Human Activities World Models. }   
Recent large-scale egocentric video datasets (e.g., Ego4D~\cite{grauman2022ego4d} and Ego-Exo4D~\cite{grauman2024ego}) have advanced human egocentric vision. However, research on comprehensive world models for this domain remains limited. 
To the best of our knowledge, UniSim~\cite{yang2023learning} is the first approach in this direction, using a video diffusion model conditioned on action labels. 
In contrast, our approach provides higher-fidelity control allowing for a greater diversity in the generated content.

\section{Uncovering the Real GEM}
\begin{figure*}[h]
    \centering
    \includegraphics[width=1.0\textwidth]{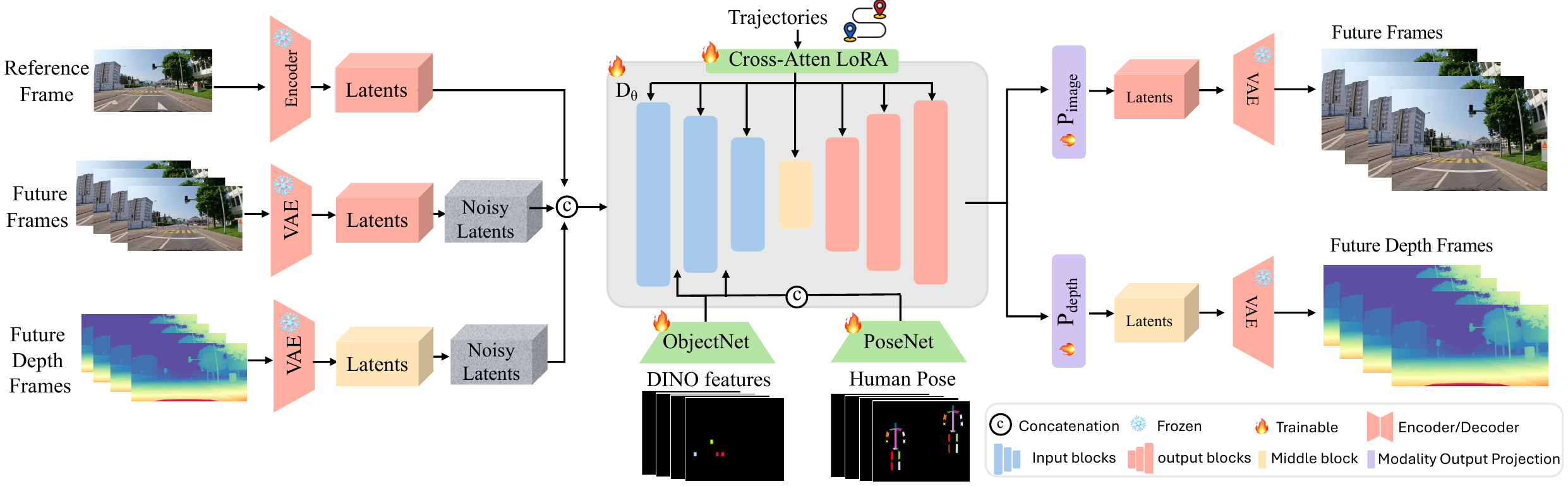}
    \caption{GEM generates two modalities by taking as inputs a reference frame and noisy latents of images and depth modalities. The denoiser network, $D_{\theta}$ is conditioned on ego trajectories, DINOv2 features and human poses. Ego-trajectories are added using a cross attention LoRA at every block of the network. DINOv2 features and human poses are added to the output of each block in the input layers of the denoiser. To handle multimodal outputs, we use different output convolution-based projection layers $P$.  }
    \label{fig:dino_control}
\end{figure*}

In this section, we present GEM's key components and capabilities. As shown in~\cref{fig:dino_control}, GEM has two output modalities—images and depth—and three control signals: ego-trajectories, DINOv2 features, and human poses. We begin with background (\cref{sec:preliminaries}), detail our control methodology (\cref{sec:method_control}), long-horizon generation (\cref{sec:method_long}), multimodal generation (\cref{sec:method-multimodal}), and training strategy (\cref{sec:method_training}).


\subsection{Preliminaries}\label{sec:preliminaries}
We cast the training of our world model as video generation. Thus, we employ the current SotA open-source image-to-video model, Stable Video Diffusion (SVD) \cite{svd}, as backbone for GEM and fine-tune it on ego-centric data. In SVD, videos are represented as sequences of $N$ RGB frames of size $H \times W$. The frames are independently encoded into the latent space of a pre-trained autoencoder, resulting in a sequence of $N$ feature maps, each having 4 channels, height $\tilde{H} = \frac{H}{8}$, and width $\tilde{W} = \frac{W}{8}$. We denote the distribution of the encoded videos from the dataset as $p_{\text{data}}(x)$, where $x \in \mathbb{R}^{N \times 4 \times \tilde{H} \times \tilde{W}}$. SVD operates within the Elucidated Diffusion Model (EDM) framework \cite{Karras2022ElucidatingTD}, where a network $D_\theta(x ; \sigma, {\cal C})$ is trained to denoise a noisy sample $x$, given the noise level $\sigma$ and conditioning variables ${\cal C}$, which may include text or video/image embeddings. In the case of SVD, ${\cal C} = \{x_0\}$ only includes the embedding of the first frame in the sequence, enabling image-to-video synthesis.

\subsection{Controlling Ego-Vision Generation} \label{sec:method_control}
We decompose the control space of our model into three main components: 1) the ego-motion, 2) the object-level control, and 3) human pose control. The first component,~\cref{sec:ego_centric_control}, allows us to specify the motion of the ego-agent through ego-trajectories.
The second component, \cref{sec:object_level_control}, facilitates object-specific control, enabling editing of scene composition and dynamics across space and time by adjusting the location of object features. This also enables insertion of new objects.
The last component, \cref{sec:method-human-pose}, enables control of pedestrian poses. 

\subsubsection{Ego-Motion Control}
\label{sec:ego_centric_control}

To control the ego-motion, we expand the set of conditioning variables in $D_{\theta}(x; \sigma, {\cal C})$ with the ego-trajectories $c_{\text{traj}}$, \ie, ${\cal C} = \{x_0, c_{\text{traj}}\}$. Ego-trajectories are metric sequences of 2D positions that quantify the motion of the ego-agent when projected to the birds-eye-view plane. Inspired by Vista~\cite{vista}, to integrate $c_{\text{traj}}$ into the network, we first embed the trajectories onto a fixed-dimensional plane and encode them using Fourier embeddings~\cite{tancik2020fourier}. Since the ego-motion control provides solely a global context and does not encode direct spatial information in the image space, 
we condition the network on $c_{\text{traj}}$ by fusing them through additional LoRA modules \cite{lora} in the cross-attention layers of the UNet backbone (see~\cref{fig:dino_control}).

\subsubsection{Object-Level Control}\label{sec:object_level_control}
For object-level control, we leverage DINOv2 tokens~\cite{dinov2}, following previous work~\cite{cage}. DINOv2 tokens are data-agnostic, abstract, and inexpensive to obtain, making them well-suited as conditioning signals. They encode high-level semantic information about objects in the scene (\eg, object categories or basic style features) and are invariant to small appearance perturbations, enabling their transfer across frames. We condition the model $D_\theta$ on a sparse set of DINOv2 tokens that specify where and when a certain object should appear in the generated video, as illustrated in \cref{fig:dino_control}. The model generates the output by inpainting the missing information both spatially and temporally. More precisely, the object-level control tokens (DINOv2 tokens) extracted from either unrelated images or the reference frame $x_0$, are inserted into zero-initialized feature maps $\{z_0, \dots, z_{N}\}, z_i \in \mathbb{R}^{N \times d \times h \times w}$ at the desired locations and times. Here, $N$ is the number of frames, $d$ is the dimension of each DINOv2 feature, and $h, w$ are its resolution. These feature maps form the object-level control $c_{\text{dino}}$ that is fed to the model, guiding it to place the objects at specific coordinates and time steps. Thus, the set of conditioning variables in $D_{\theta}$ becomes ${\cal C} = \{x_0, c_{\text{traj}}, c_{\text{dino}}\}$.

The training of the object-level control is done in an unsupervised manner. During training, we randomly sample $k$ frames $\{x_{t_1}, \dots, x_{t_k}\}$ from a given video $x \sim p_{\text{data}}(x)$. We then process the original frames with DINOv2, and extract the corresponding dense feature maps $\{z_{t_1}, \dots, z_{t_k}\}$, where $z_{t_i} \in \mathbb{R}^{d \times h \times w}$. From each of these feature maps, we randomly mask all but $m \sim U[0, M]$ tokens, where $M$ is a hyperparameter that we set to 32 in our experiments. The masked feature maps are then padded with zero maps to match the original frame count. Thus, we obtain $c_\text{dino} = \{z^{\text{masked}}_{t_1}, \dots, z^{\text{masked}}_{t_k}\}^{\text{pad}}$. By employing this randomized approach, we foster learning of both the spatial composition and the temporal dynamics of the scene.

\begin{figure}[t]
    \centering
    \includegraphics[width=0.8\linewidth]{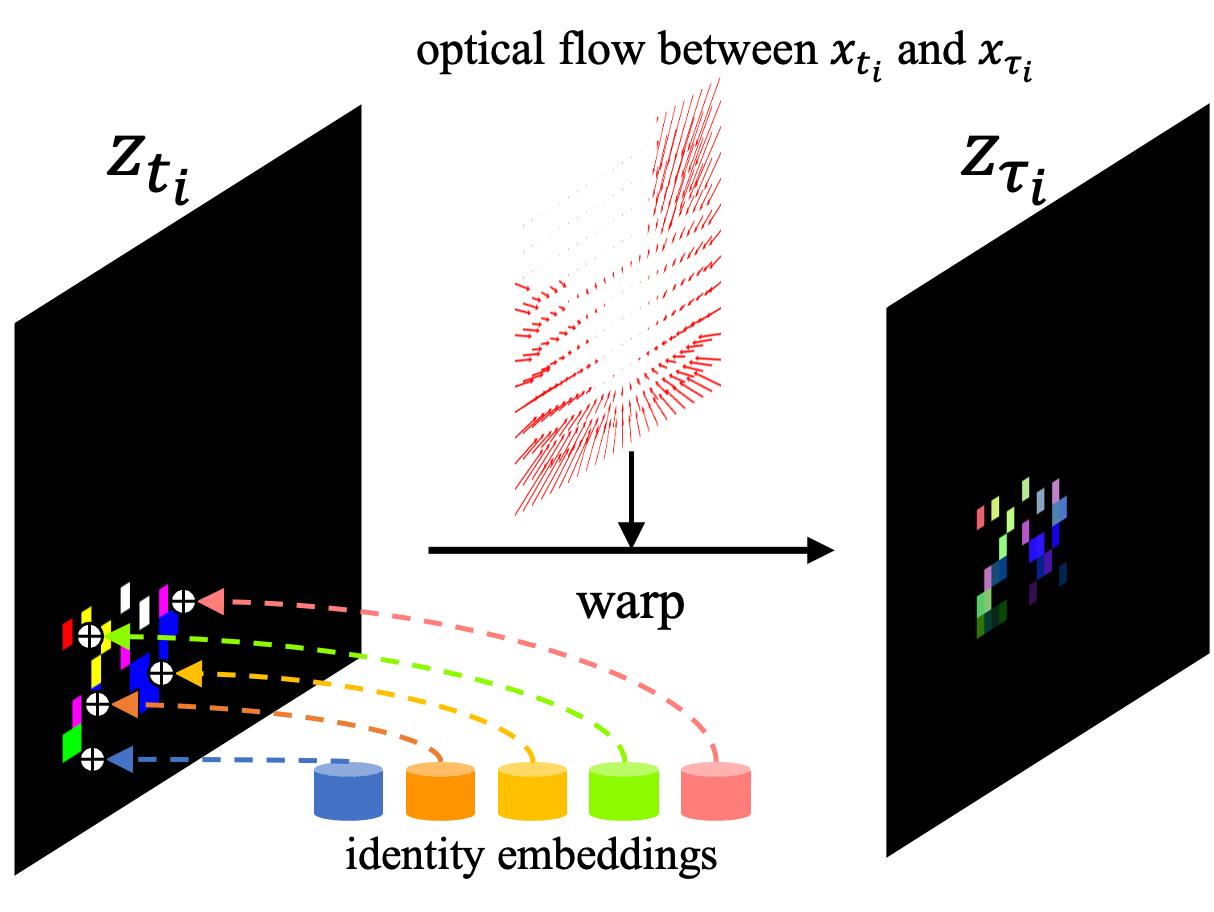}
    \caption{During training the sparse DINOv2 features from frame $t_i$ are translated to frame $\tau_i$ using the corresponding optical flow.}
    \label{fig:id}
\end{figure}

\noindent\textbf{Identity Embeddings.}
One challenge in the object-level control using DINOv2 features arises when the inserted tokens are both spatially and feature-wise similar to the visual features of objects already present in the reference frame. This creates an ambiguity between moving an existing object or inserting a new one.
To address this, we propose using a learned identity embedding to associate individual tokens over time. This approach, visualized in \cref{fig:id}, 
involves adding the same identity embedding to the control tokens representing the same moving entity across different time steps. More specifically, we start with $\{z^{\text{masked}}_{t_1}, \dots, z^{\text{masked}}_{t_k}\}$, as before, and add individual learned identity embeddings $\text{ID}_\phi: \{1,... L\}\rightarrow \mathbb{R}^d$ to the nonzero tokens in each map. Here $L$ is chosen to be large enough to ensure that different tokens from the same feature map do not receive the same identity embedding. For each feature map, we then sample a target time $\tau_i > t_i$ and translate the tokens from $z_{t_i}^{\text{masked+ID}}$ to $z_{\tau_i}$ using the optical flow between frames $x_{t_i}$ and $x_{\tau_i}$, as shown in \cref{fig:id}. At inference, we can disambiguate the generation by using the same identity embeddings in the reference and in the target frames to guide the model towards moving the underlying object instead of introducing a new object at the desired location.

\noindent\textbf{Conditioning Technique.}
In contrast to the ego-motion, the object-level control encodes fine-grained details about the scene composition. This influences the design of the conditioning technique as it is now necessary to incorporate spatial information. We start by processing the sequence of sparse DINOv2 feature maps using a network with a similar architecture to the input blocks of the denoising UNet. We call this network \textit{ObjectNet}. ObjectNet is meant to capture and inpaint both the spatial and the temporal information in the sparse DINOv2 feature maps. Similarly to other works~\cite{mimic}, the encoded tokens are directly added to the outputs of the UNet's input blocks, as depicted in~\cref{fig:dino_control}. Empirically, we observe that this technique for the object-level control outperforms feeding the DINOv2 tokens through cross-attention layers, as was done for the ego-motion control. Moreover, the use of ObjectNet acts as a transition layer, bridging the domain gap between DINOv2 and the UNet's internal feature space, and outperforms fusing pure DINOv2 feature maps into the denoiser.

\subsubsection{Human Pose Control} \label{sec:method-human-pose}
We find that the aforementioned object-level control performs well for objects with few moving parts. However, generating accurate representations of humans remains a challenge for the model. Nonetheless, to facilitate safe navigation and human-robot interaction, it is crucial to model humans accurately. Therefore, we extend the object-level control with a human pose component, \ie ${\cal C} = \{x_0, c_{\text{traj}}, c_{\text{dino}}, c_{\text{pose}}\}$. To condition the model $D_\theta$ on the extracted human poses, we follow previous techniques for generating human motion~\cite{mimic}; we draw the skeletons on an empty image plane and pass it through a CNN, PoseNet, to embed the spatial information. We then add the human pose feature maps to the network features, $D_{\theta}$, in a similar way to the object-level controls as shown in~\cref{fig:dino_control}.



\subsection{Stable Long Video Generation}\label{sec:method_long}

\begin{figure}[t]
    \centering
    \includegraphics[width=0.49\textwidth]{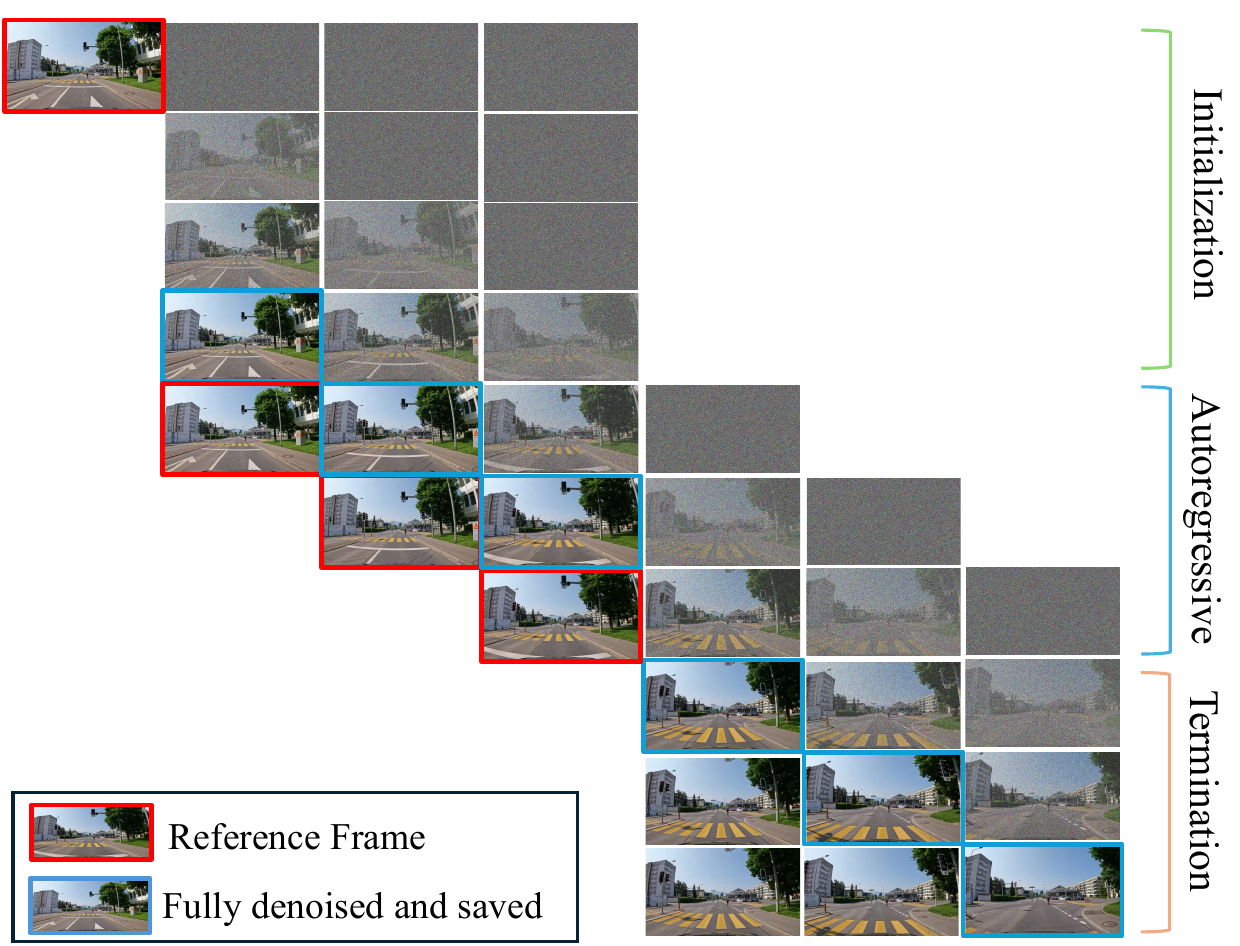}
    \caption{Visualization of our dynamic autoregressive sampling noise schedule for denoising 6 frames in total with a window size of 3 frames and 3 sampling steps.}
    \label{fig:sampling_methods}
\end{figure}

Generating long videos beyond the training horizon is a challenging task for diffusion models~\cite{zhang2024mimicmotion,chen2023seine, chen2024diffusion,xie2024progressive}. A simple approach is to generate sequential short clips with overlapping frames. 
However, this causes temporal discontinuities and abrupt scene changes~\cite{xie2024progressive}.  Inspired by recent works~\cite{chen2024diffusion,xie2024progressive}, we introduce progressive denoising and autoregressive sampling, using a per-frame noise schedule to reinforce causal relationships between consecutive frames.



\noindent\paragraph{Autoregressive Sampling.} The goal of sampling is to autoregressively denoise all frames over a long horizon. 
To this end, we adopt a dynamic per-frame noise schedule, as illustrated in~\cref{fig:sampling_methods}.
The schedule has three phases: initialization, autoregressive, and termination. Initially, the schedule controls the noise levels of each frame so that the denoising of frame $i$ only starts after the denoising of frame $i-1$ has been initiated. This enables the denoising of each frame to benefit from some cleaner information in its preceding frames. Once a frame is fully denoised, it is saved and replaces the current reference frame. 
At this point, the autoregressive phase starts where at each step, a fully denoised frame is removed and a new noisy frame is appended. 
This process continues until only $N$ frames still need to be denoised, signaling the start of termination stage. At this stage, no new frames are added; the fully denoised frames are saved
(see algorithm in supplementary material). \\

\noindent\textbf{Training Noise Schedule.} 
To support the inference with the proposed custom noise schedule, we modify the training noise distribution in the following way. We first sample a random noise level $\log(\sigma) \sim \mathcal{N}(p_\text{mean}, p_\text{std})$. Using SVD's noise-to-time step mapping, we compute the corresponding denoising time step, $t_\text{intercept}$. Next, we sample a random shift $t_\text{shift} \sim \text{Beta}(\alpha, \beta)$, where $\alpha$ and $\beta$ are selected to favor lower shift values. The per-frame time steps are then calculated as $t_\text{intercept} - (\frac{i}{N - 1} - t_\text{shift})$ for $i \in \{0, \dots, N - 1\}$, ensuring a consistent noise increase over the frame axis.
To add variability, we add small random noise to the time steps, which are subsequently converted back to $\sigma$ values. This approach keeps the essential information within the attention window for the autoregressive component.

\subsection{Multimodal Generation } \label{sec:method-multimodal} 

We incorporate depth as an additional generated modality to leverage its rich spatial information, proven to enhance tasks such as scene perception, planning, object localization, and more~\cite{wang2019pseudo,achtelik2009stereo}.
By generating depth alongside RGB images, GEM can generate spatial information alongside the structural context of the scene.
To encode and decode depth, we use the same VAE used for images, following~\cite{hu2024depthcrafter}, which shows that the pretrained VAE of SVD has negligible reconstruction error on depth images. 
We concatenate both modalities at the input and introduce an output convolution projection layer ($P_{\text{depth}}$) to the denoising network to predict the noise for the depth. $D_{\theta}$ simultaneously denoises both inputs ensuring consistency between both modalities. Therefore,  the final denoiser is $D_\theta(x, x_\text{depth}; \sigma, \{x_0, c_\text{traj}, c_\text{dino}, c_\text{pose}\})$.




\subsection{Training Strategy} \label{sec:method_training}
For efficiency, we divide our training into two distinct stages, where the first one focuses on learning new control signals, and the second stage emphasizes high-resolution generation. We begin with the pre-trained SVD~\cite{svd} and initially fine-tune it on low-resolution videos (320$\times$576) using all added control signals and modalities. In the second stage, the training continues in the same way, but at a higher resolution (576$\times$1024). As detailed in~\cref{sec:data-curation}, we apply data filtering to improve diversity and quality during both stages. For more details, refer to the supplementary material.

\section{Dataset Preparation} \label{sec:method_data}
\vspace{-1mm}
We combine various open source datasets across different domains presented in~\cref{tab:datasets}. 
We use 3211 hours of driving, 1000 hours of human egocentric videos,  and 27.4 hours of drone footage that we collected from YouTube.\\

\noindent\textbf{Data Curation.} \label{sec:data-curation}
To achieve precise control over object movements, the training data must include (1) diverse interactions and dynamics, (2) fine-grained object details. We curate the dataset by removing low-quality and low-motion sequences, segmenting videos into 2.5-second clips, and applying two types of filters: \emph{quality} and \emph{diversity}. Quality filtering excludes clips with poor camera quality or high blur using aesthetic scores from the LAION dataset~\cite{aesthetic} and PIQE metrics~\cite{piqe}, similar to~\cite{moviegen}. Diversity filtering assesses motion diversity via optical flow, similar to~\cite{moviegen, Girdhar2023EmuVF}, and semantic variation using DINO feature encodings~\cite{dinov2}. Clips with low intra-clip diversity or high cross-clip similarity are excluded to balance motion and content (more details in the supplementary material).\\

\noindent\textbf{Pseudo-labeling.}\label{sec:pseudo}
Given the scarcity of labeled datasets, we pseudo-label all the data with depth information, ego trajectories, and human skeletons. 
We generate metric depth using Depth Anything V2~\cite{yang2024depth} for trajectory labeling and geometric understanding. Ego trajectories are estimated with GeoCalib~\cite{veicht2024geocalib} for intrinsics followed by DroidSLAM~\cite{teed2021droid} for RGB-D SLAM, with pseudo-depth resolving scale ambiguity. Finally, human skeletons are labeled using DWPose~\cite{yang2023effective} for efficient and high-quality pose estimation. More details are provided in the supplementary material.

\begin{table}[t]
    \centering
    \resizebox{\columnwidth}{!}{%
    \begin{tabular}{cccccc}
        \toprule
        Domain & Dataset & Hours & \begin{tabular}[c]{@{}c@{}}Front-View \\ Frames\end{tabular}   & \begin{tabular}[c]{@{}c@{}}Diversity \\ Cities\end{tabular} &\begin{tabular}[c]{@{}c@{}}GT \\ Traj\end{tabular}     \\
        \hline
        \multirow{17}{*}{\textbf{Driving}}& OpenDV~\cite{genad}& 1747 & 60.2M & $>$244 &  \\
        & BDD~\cite{bdd}& 1000 & 100M & 1 &  \\
        & Nuscenes~\cite{nuscenes} & 5.5 & 241k &2 & \checkmark \\
        & Driving Dojo~\cite{drivingdojo} & 150 & - & 9 & \checkmark \\
        & Honda HDD~\cite{hondahdd} & 104 & 1.1M & 1 & \\
        & Honda HAD~\cite{kim2019CVPR} & 30 & - & 1 & \\
        & Drive360~\cite{hecker2018end} & 55 & - & - & \\
        & D${}^2$City~\cite{che2019d} & 100 & 700k & 5 & \\
        & DoTA\accident~\cite{yao2020and} & 13.9 & 504k & - & \\
        & CarCrashDataset\accident~\cite{bao2020uncertainty} & 6.25 & - & - & \\
        \cline{2-6}
        & \textbf{Total} & 3211 &$>$162.4M & $>$244 & \\
        \hline
        \textbf{Human} & EgoExo4D~\cite{grauman2024ego} & 1000 & 108M & - & \checkmark\\
        \hline
        \textbf{Drone} & self-collected & 27.4 & 961k & - & \\
        \bottomrule
    \end{tabular}}
    \caption{Overview of the ego video datasets used during training, totaling more than 4000 hours of training data. {\tiny \accident} denotes accident-focused datasets.}
    \label{tab:datasets}
\end{table}

\section{Experiments } \label{experiments}
\begin{figure*}[t]
    \centering
    \includegraphics[width=\textwidth]{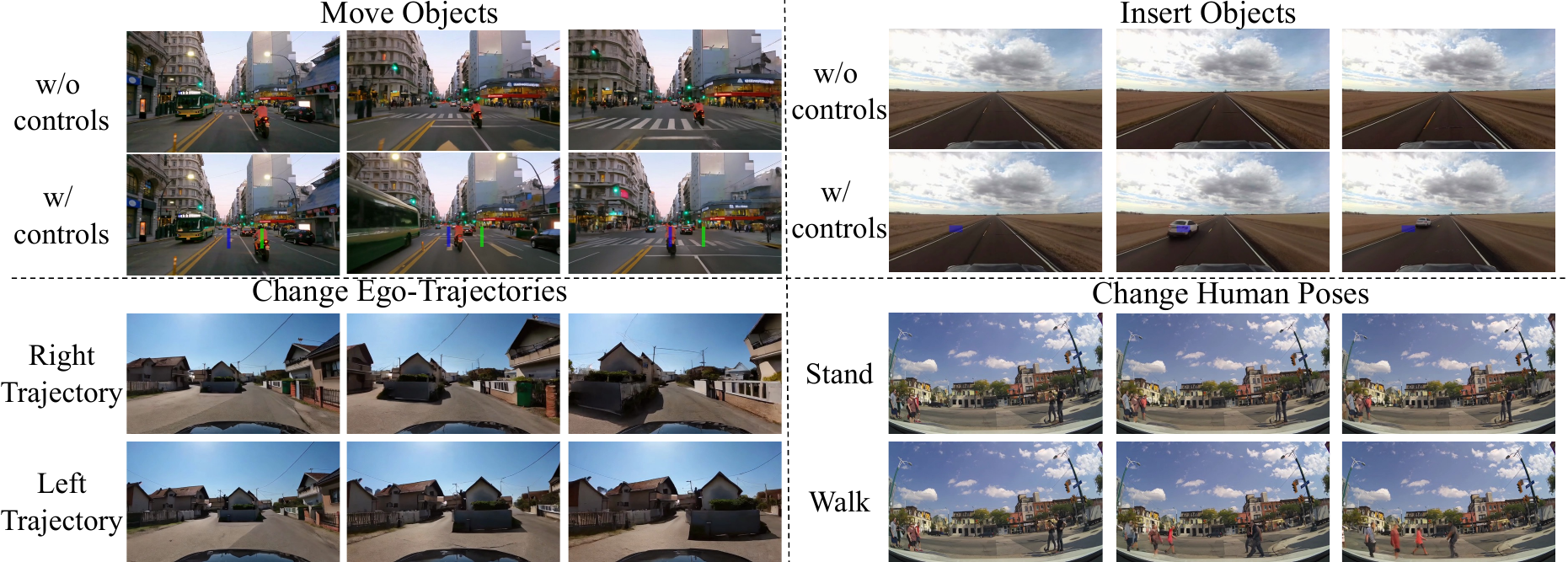}
    \caption{Qualitative results for GEM's controllability. GEM can flexibly move objects (top-left), insert new objects (top-right), change ego trajectories (bottom-left) and change human poses (bottom-right). Refer to our \href{https://vita-epfl.github.io/GEM.github.io/}{website} for more videos.}
    \label{fig:results_image}
\end{figure*}

In this section, we conduct experiments to evaluate our model based on quality and controllability.  We start by introducing our metrics in~\cref{sec:method_eval}. Then, we outline our experiments for quality evaluation followed by controllability evaluation.  
The experiments are conducted on Nuscenes' validation set and a randomly sampled subset of equal number of videos from the OpenDV's validation set. We use Vista~\cite{vista} as a baseline, as its architecture aligns closely with ours, making it the most comparable model to GEM. We additionally show qualitative results in~\cref{fig:results_image}.


\subsection{Evaluation Metrics} \label{sec:method_eval}
We outline the metrics utilized to assess various aspects of our world model.\\

\noindent\textbf{Video Quality.} We evaluate the generation quality of our world model using standard metrics: Frechet Inception Distance (FID) \cite{Heusel2017GANsTB} and Frechet Video Distance (FVD)~\cite{Unterthiner2018TowardsAG}. \\

\noindent\textbf{Ego Motion.} To evaluate the ego-motion control, we estimate the ego-motion, $\hat{p}$, from the generated videos using our pseudo-labeling technique (\cref{sec:pseudo}). We use the Average Displacement Error (ADE) to compare the generated trajectory against the ground truth trajectories. If a dataset has no ground truth labels, we estimate the pseudo-ground truth using the same pipeline (\cref{sec:pseudo}).\\

\noindent\textbf{Control of Object Manipulation (COM).} 
To evaluate the object-level control, we use YOLOv11~\cite{yolov11} to detect and track the objects through frames. We extract the bounding boxes of the largest vehicle in the scene and compare its bounding boxes across the frames in the generated and ground-truth videos. We then calculate the absolute difference in pixels between the centers of the bounding boxes as follows: $\text{COM}=|\text{BBox}_\text{gen} - \text{BBox}_\text{GT}|$. \\

\noindent \textbf{Human Pose.}
For human poses, we select video clips that contain at least five pedestrians and extract their joints using DWPose~\cite{yang2023effective}. We use the COCO toolkit to evaluate based on 17 keypoints; we calculate Average Precision (AP) of poses extracted from the generated videos against those from the ground truth videos. We compare AP of poses in unconditional versus conditional generation to evaluate the effectiveness of the control technique. \\

\noindent\textbf{Depth Evaluation.}
We compare our generations to the pseudo-labels. The evaluation is based on two commonly used metrics~\cite{yang2024depth,hu2024depthcrafter, depthanything}: Absolute Relative Error (AbsRel) defined as $\frac{|\hat{d} - d|}{d}$,  and $\delta$ defined as percentage of $\max\left(\frac{d}{\hat{d}}, \frac{\hat{d}}{d}\right) < 1.25$. Results in supplementary material.

\subsection{Comparisons of Generation Quality}
\noindent\textbf{Training-Horizon Generation Quality.}
\cref{tab:performance_comparison} compares GEM's generation quality with existing autonomous driving world models based on the standard quality metrics, FID and FVD. As shown in~\cref{tab:performance_comparison}, GEM outperforms Vista on FVD results for both datasets and achieves competitive FID results for OpenDV. While GEM's FID results on Nuscenes are marginally lower than Vista's, it is likely due to Vista's fine-tuning on Nuscenes during the final training stage. Notably, GEM achieves these comparable results despite being trained on a dataset curated specifically to enhance controllability performance rather than visual quality. \\

\begin{table}[t]
    \centering
    \resizebox{\columnwidth}{!}{%
    \begin{tabular}{ccccc}
        \toprule
        & \multicolumn{2}{c}{Nuscenes} & \multicolumn{2}{c}{OpenDV} \\
        & FID $\downarrow$ & FVD $\downarrow$ & FID $\downarrow$ & FVD $\downarrow$ \\ \midrule
        DriveGAN~\cite{kim2021drivegan} & 73.4 & 502.3 & - & - \\
        DriveDreamer~\cite{drivedreamer} & 14.9 & 340.8 & - & - \\
        DriveDreamer-2~\cite{drivedreamer2} & 25.0 & 105.1 & - & - \\
        WoVoGen~\cite{wovogen} & 27.6 & 417.7 & - & -\\
        Drive-WM~\cite{drivewm}& 15.8 & 122.7 & - & -\\
        GenAD \cite{genad} & 15.4 & 184.0 & - & - \\
        \hline
        Vista \cite{vista} & 6.6$^\star$ & 167.7$^\star$& 5.5$^\star$& 163$^\star$\\
         GEM (Ours) & 10.5 & 158.5&6.3  &131\\\bottomrule
    \end{tabular}}
    {\footnotesize $^\star$ our reproduced results using the official code from Vista~\cite{vista}.}
    \caption{Quality comparison of generations on Nuscenes and OpenDV datasets}
    \label{tab:performance_comparison}
\end{table}

\begin{figure}[t]
    \centering
    \includegraphics[width=0.42\textwidth]{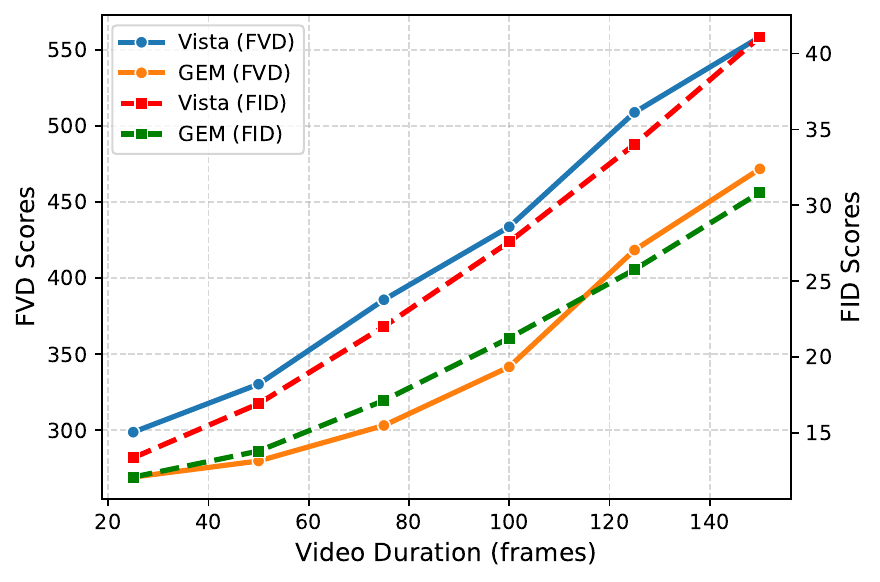}
    \caption{FVD and FID comparison for the long generations of GEM and Vista~\cite{vista}.}
    \label{fig:quality-long}
\end{figure}

\noindent\textbf{Long-Horizon Generation Quality.} 
We quantitatively evaluate long-generation quality by randomly generating 500 videos of 150 frames using GEM’s autoregressive sampler and Vista’s triangular sampler (window size: 25 frames, 3-frame overlap). Due to the computational cost, we limited the experiment to 500 randomly generated videos. We calculate FID and FVD on video lengths of 25, 50, 75, 100, 125, and 150 frames. Results (\cref{fig:quality-long}) show GEM outperforming Vista, with consistently lower FVD and FID scores across all durations, reflecting better temporal consistency and quality. 
This demonstrates GEM’s superior sampling approach for long-horizon videos. 

\subsection{Human Evaluation}

\begin{figure}[t]
    \centering
    \includegraphics[width=0.49\textwidth]{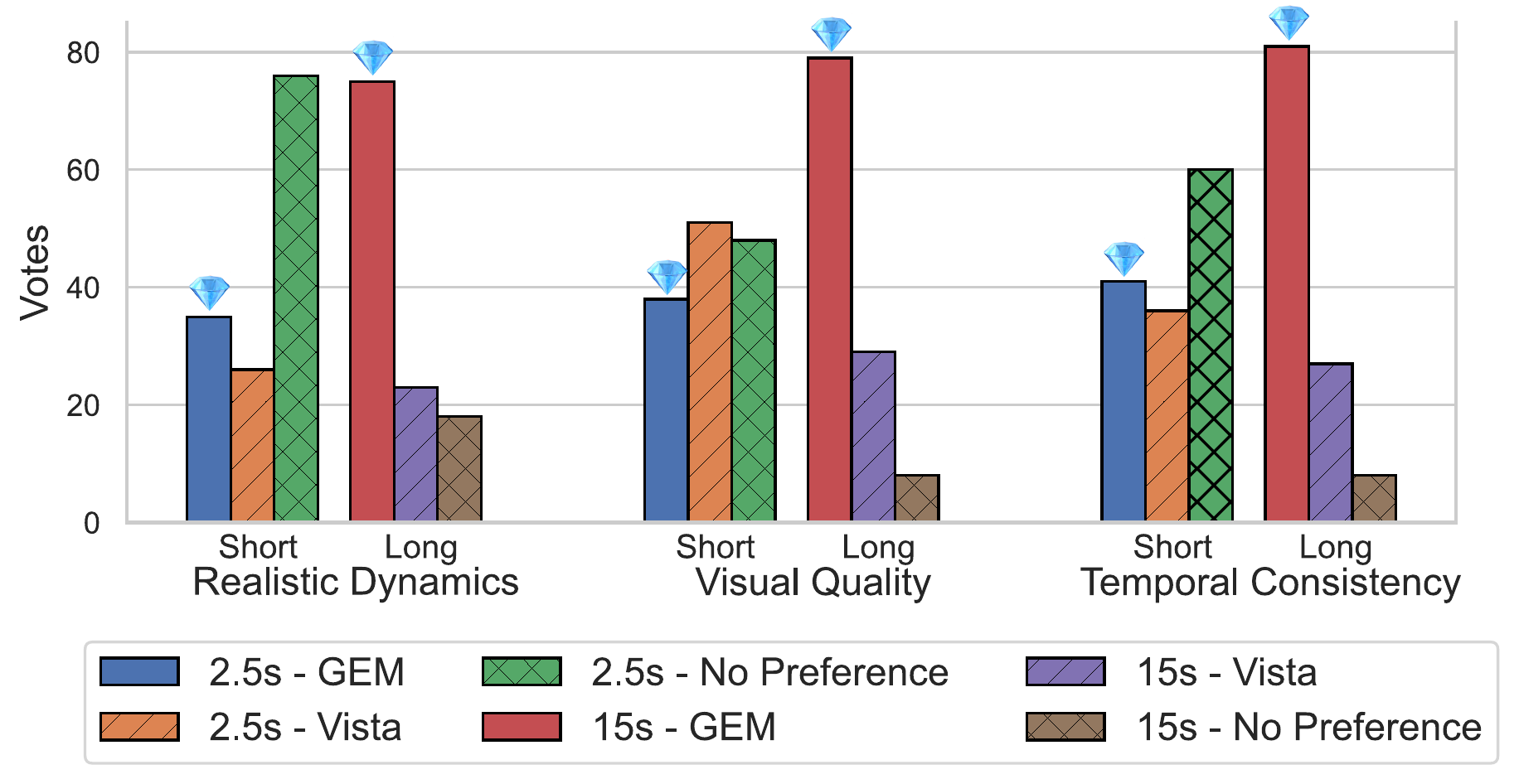}
    \caption{Human evaluation results on short (2.5s) and long (15s) videos. The gem on top denotes GEM.}
    \label{fig:human-eval-combined}
\end{figure}

To address the limitations of FVD in evaluating video perceptual quality, particularly for dynamic scenes~\cite{blattmann2023align, brooks2022generating}, we conduct a human evaluation. Participants performed pairwise comparisons between GEM and Vista’s unconditional video generations. Using a random selector, we sampled 50 2.5s and 50 15s videos from each model, along with 50 videos of each length from the OpenDV validation set. Each participant evaluated 20 randomly selected videos, equally split between short and long generations, focusing on realistic dynamics, visual quality, and temporal consistency. Results, based on 116 responses, are shown in \cref{fig:human-eval-combined}.


For short videos, most participants found minimal differences between GEM and Vista, with 76, 48, and 60 noting no distinction in realism, visual quality, and temporal consistency, respectively. This suggests that short generations are generally perceived as highly similar.


For long videos, GEM was strongly preferred: 75 vs. 23 votes for realism, 79 vs. 29 for visual quality, and 81 vs. 27 for temporal consistency. These results indicate that GEM’s long generations are perceived as superior in realism, visual quality, and temporal consistency as per human judgment.

\vspace{-2mm}

\subsection{Comparisons of Controllability}\label{sec:comparison_controllability}
In this section, we evaluate controllability of ego motion, object motion and human pose. We compare our conditional generation against the unconditional one to evaluate the effectiveness of our control strategy. \\

\noindent\textbf{Ego-Motion}~\cref{tab:controllability_comparison} reports the ADE results of generated trajectories for the Nuscenes dataset and OpenDV.
Results on Nuscenes show slight improvement in our controlled generation.
This is due to the fact that in most cases, the ego-motion is obvious in the 2.5s window (\eg, moving straight). Therefore, we manually choose a subset of 50 videos, Nuscenes$^{sub}$, where ambiguity is present (\eg, multiple paths ahead). Results on Nuscenes$^{sub}$ in~\cref{tab:controllability_comparison} show 18\% improvement of the conditional generation, demonstrating effective ego-motion controllability. We additionally provide qualitative examples on our \href{https://vita-epfl.github.io/GEM.github.io/}{website}.  \\


    

\begin{table*}[t]
    \centering
 \resizebox{\textwidth}{!}{%
    \begin{tabular}{lccc|cc|cc}
        \toprule
        & \multicolumn{3}{c|}{Ego-Motion Controllability} &  \multicolumn{2}{c|}{Object Motion Controllability} &  \multicolumn{2}{c}{Human Pose Controllability} \\
        & \multicolumn{1}{c}{Nuscenes} & Nuscenes$^{sub}$ & OpendDV & \multicolumn{1}{c}{Nuscenes} & \multicolumn{1}{c|}{OpenDV} & \multicolumn{2}{c}{OpenDV}  \\
        & ADE $\downarrow$  & ADE $\downarrow$ & ADE $\downarrow$ & COM $\downarrow$  & COM $\downarrow$ & \begin{tabular}[c]{@{}c@{}} AP@IoU=.5  $\uparrow$ \\ area=all \end{tabular}   & \begin{tabular}[c]{@{}c@{}} AP@IoU=.5:.95  $\uparrow$ \\ area=large \end{tabular}\\ \midrule
        GEM  \textit{w/o controls} &  3.24& 3.59 & 5.39 & 38.8 & 55.2 & 0.00 & 0.00\\
        GEM  \textit{w/ controls} & \textbf{3.07} & \textbf{2.85} & \textbf{3.47}  &  \textbf{12.2} & \textbf{11.5}  & \textbf{0.12} & \textbf{0.12} \\\bottomrule
    \end{tabular}} \\
    \caption{Controllability evaluation of ego-motion, object motion and human pose.  Nuscenes$^{sub}$ denotes the subset of Nuscenes with ego-motion ambiguity. }
    \label{tab:controllability_comparison}
    
\end{table*}

    

    

\noindent\textbf{Object-Motion}
~\cref{tab:controllability_comparison} reports the results of object manipulation controllability based COM introduced in~\cref{sec:method_eval}. 
The results show that our conditional generation consistently outperforms the unconditional one by 68.8\% on Nuscenes and 79\% on OpenDV. The performance gain highlights the effectiveness of our control strategy for moving objects in a scene. \\

    

    


    
\noindent \textbf{Human Pose}
We report two types of AP metrics: AP based on (1) loose Intersection over Union (IoU) of at least 50\% on all sizes of skeletons, (2) strict IoU (50\% to 95\%) on large skeletons. \cref{tab:controllability_comparison} presents the results, highlighting the control effectiveness particularly for large human poses. 

\section{Conclusion}
We introduced GEM, a multimodal world model for ego-vision tasks, capable of generating videos in environments with complex dynamics. By leveraging egocentric trajectories, DINO features, and human skeletons, GEM enables precise control over ego-motion, object movements, as well as humans. Its multimodal features, including image and depth frame generation, provide both rich semantic and spatial context. Our evaluation results have highlighted GEM's effectiveness, with conditional generation significantly surpassing unconditional generation for ego trajectories, object motion, and modeling humans. While GEM advances the state of the art in controllable ego-vision world models, it is not without limitations. Notably, while GEM demonstrates strong performance in long generations, further improvements are needed to enhance their quality and consistency over extended sequences. Despite these limitations, we hope GEM will serve in the future as a foundation for adaptable and controllable world models. 

\noindent \textbf{Acknowledgments.} This work was supported as part of the Swiss AI Initiative by a grant from the Swiss National Supercomputing Centre (CSCS) under project ID a03 on Alps. Sebastian Stapf and Aram Davtyan have been supported by SNSF Projects 200020-188690 and 200021-228098. Ahmad Rahimi has been supported by Hasler Foundation under the Responsible AI program.

{
    \small
    \bibliographystyle{ieeenat_fullname}
    \bibliography{main}
}

\clearpage
\setcounter{page}{1}
\maketitlesupplementary



\textbf{Q1: Will you share the code and dataset publicly? } 

All codes and model checkpoints are publicly available at our \href{https://github.com/vita-epfl/GEM}{GitHub repo} including all scripts used for pseudo-labeling for reproducibility. We additionally intend to release the pseudo-labels of the dataset.


\textbf{Q2: How accurate are the trajectories pseudo-labeling?} 

We evaluate our pseudo-labeling pipeline which consists of calibration, depth estimation, and finally SLAM, on Nuscenes. We show the results in~\cref{sec:pseudo-ego-supp} based on the Average Displacement Error (ADE) which are 0.48 meters with scale compensation and 1.68 meters without scale compensation. Although both these numbers are acceptable, this difference marks the inherent error in the monocular depth estimator model we use. We find the accuracy of the trajectories high enough to use them as control signals for GEM.  

\textbf{Q3: Are the depth and images synced?}

By observing the generations, it is evident that the modalities are aligned. This alignment can be attributed to several factors: (1) both modalities are encoded using the same model, ensuring similar latent representations and preserving spatial correspondences; (2) they are processed simultaneously through the network, allowing it to learn and model relationships between the modalities; and (3) the same sampler is used for both. These factors collectively contribute to the alignment observed between the two modalities. Refer to our \href{https://vita-epfl.github.io/GEM.github.io/}{website} to see many examples on both modalities.

\textbf{Q4: Why are different controls added with different techniques?}

For ego motion, we empirically find that it is sufficient to incorporate the trajectories using additional cross-attention layers. However, this was not the case for the other controls that did not show similar effectiveness when added through additional cross attention layers. For object and human pose controls, where fine-grained details in the encoding of the scene composition is needed, it is necessary to incorporate spatial information. Therefore, we use specific networks to project these controls and add these features to the output of the backbone's input blocks. 

\textbf{Q5: What's the strategy used to evaluate the different controls? }

Our evaluation strategy for all control techniques involves applying the control, detecting it in both the generated video and the ground truth, and comparing the results using a specific metric.  For example, for ego motion control, we apply the trajectory control and get the generated video. For datasets without ground-truth labels, we use our pseudo-labeling pipeline to detect the trajectories in both the generated video and the ground-truth video. We then use Average Displacement Error (ADE) for comparison. We use similar technique with other controls, each having a specific evaluation metric. If no similar world model can perform the same control strategy, we compare our conditional generations against the unconditional ones to show the effectiveness of our control. 

\textbf{Q6: Is data curation needed? What’s the motivation behind it?}

We incorporate a large amount of uncurated data into the dataset. Many samples suffer from poor camera distortions, extreme blurriness, or are completely black. Therefore, a quality filtering step is necessary. 
Furthermore, the dataset contains numerous videos with minimal activity, \textit{e.g.}, long highway drives. To enhance training efficiency, we filter the dataset based on diverse scene characteristics. 
Furthermore, to achieve precise control over object movements within a scene, the training data must: (1) include diverse interactions and dynamics, and (2) capture fine-grained details of the objects. This ensures the training process supports accurate control mechanisms while maintaining efficiency.

\section{Method}
\subsection{Pseudo-labeling}

\textbf{Depth.} 
We generate depth information for (1) trajectory pseudo-labeling and (2) generating the spatial information of the scene. For depth estimation, we utilize the metric version of Depth Anything V2-Small~\cite{yang2024depth}, a state-of-the-art depth estimator known for its accuracy on the KITTI dataset and per-frame consistency.

\noindent\textbf{Ego-trajectories.} \label{sec:pseudo-ego-supp}
To estimate ego-trajectories, we first determine the camera's intrinsic parameters with GeoCalib~\cite{veicht2024geocalib}, using a pinhole camera model. For videos with radial distortion, we empirically find that radial camera calibration yields improved results. Using the estimated intrinsics and the RGB-depth output from Depth Anything V2, we then apply DroidSlam~\cite{teed2021droid}, an RGB-D SLAM algorithm. The use of metric depth is crucial to help with scale ambiguities. The output of the SLAM algorithm consists of a sequence of camera-to-world matrices {\( A_i = \begin{bmatrix} R & T \\ 0 & 1 \end{bmatrix} \) for \( i \in \{1, \dots, N\} \)}. 
For driving scenes, we extract X and Z displacements to have Bird-eye-view trajectories, and for ego-centric domains, we include Y displacement and the rotations in the Ortho6D format 
\cite{zhou2019continuity}.

We evaluate our trajectory pipeline on the NuScenes dataset using ground truth trajectories as a benchmark. As shown in~\cref{tab:traj_nuscenes}, the Average Displacement Error (ADE) is 1.64 m when scale is not compensated, relying solely on the depth pseudo-labels to guide the scale. However, when we compensate for the scale using ground truth labels, the ADE is reduced to 0.48 m. This result highlights the potential value of improving depth annotations, as better depth quality could further enhance trajectory accuracy. Despite this limitation, our pseudo-labeled trajectories are sufficiently accurate to guide the model in controlling the motion of the ego vehicle. In our use case, the primary requirement is for the trajectories to approximate the motion of the ego agent closely enough to enable the model to generalize and control the vehicle in new scenarios effectively.

\begin{table}[h]
    \centering

    \begin{tabular}{cc}
        \toprule
        & \multicolumn{1}{c}{Nuscenes} \\
        & ADE (m) $\downarrow$ \\ \midrule
        With Scale Compensation& 0.48  \\
        Without Scale Compensation& 1.63 \\
        \bottomrule
    \end{tabular} \\
    \caption{Trajectory pipeline evaluation on Nuscenes }
    \label{tab:traj_nuscenes}
    
\end{table}

\textbf{Human Pose.} We generate human poses using DWPose~\cite{yang2023effective}. The annotation of each human is 17 keypoints  describing all the body joints. 

\subsection{Sampling Algorithm} 
\cref{alg:dynamic_pyramid} introduces the sampling technique used with dynamic noise schedule. The scheduling matrix $S$ governs the progression of noise levels across frames, with values adjusted based on the temporal relationship between the scheduling index and frame indices. The noise schedule dynamically adjusts to three different phases: initialization, autoregressive and termination. The initialisation starts denoising the frames at different timesteps till the first frame is fully denoised and the last frame just started a few denoising steps. Autoregressive phase gets a fully denoised frame at each step which gets saved and a new column is appended for a new frame. Once we cannot append any more frames, termination starts and the rest of the frames are progressively denoised without appending new ones.

\begin{algorithm}[t]
\caption{Sampling with Dynamic Noise Schedule}
\label{alg:dynamic_pyramid}
\begin{algorithmic}[1]
\Require Initial noisy frames $x \in \mathbb{R}^{F \times H \times W}$, noise schedule $\{\sigma_t\}_{t=1}^{T}$, chunk size $C$.
\State Compute the scheduling matrix $S \in \mathbb{R}^{H \times F}$:
\[
S(m, t) = 
\begin{cases} 
\sigma_0 & \text{if } t > m \\
\sigma_{m - t} & \text{if } m - t < |\sigma| \\
\sigma_{|\sigma| - 1} & \text{otherwise}
\end{cases}
\]
\State i = 0 , f = 0 \Comment{Set row index and frame index to 0}
\While{frames remain to be denoised}

        \State Apply denoise step
        \[
        \begin{aligned}
        x[f:f+H] &= x[f:f+H] + \Delta_t, \\
        \Delta_t &= \text{DenoiseStep}(x, S[i, :] ,S[i+1, :] )
        \end{aligned}
        \]
        
        \If{$f = 0$ (first iteration)}
            \State \textbf{Initialization Phase:} 
            \State Frames $x[f:f+H]$ begin denoising with scheduling matrix $S$.
        \ElsIf{$F - f > C$ (frames are appended)}
            \State \textbf{Autoregressive Phase:}
            \State Update scheduling matrix \( S \) by shifting columns and adding a new column:
            \[
            S = \text{ShiftLeft}(S), \quad S[:, -1] = \{\sigma_t\}_{t=1}^{T}
            \]
        \Else
            \State \textbf{Termination Phase:}
            \State Stop appending new frames. Continue denoising with the remaining columns of \( S \):
            \[
            S = S[:, :F - f]
            \]
        \EndIf
    \If{fully denoised frame}
        \State Save the fully denoised frame $x[f]$.
        \State Increment $f = f + 1$.
    \EndIf
    \State i = i+1
\EndWhile

\State \textbf{Return} fully denoised frames $x_{\text{denoised}}$.
\end{algorithmic}
\end{algorithm}

\subsubsection{Time complexity}
Here we discuss the time complexity of our sampling algorithm. Assume we want to generate a video of $F$ frames, each denoised in $d = 25k$ steps. In the initialization phase, frame $1\leq i \leq 25$ gets denoised $(25-i+1)k$ times, requiring $25k$ forward passes of the model. After this phase, the first frame is clean and the 25th frame is denoised for $k$ steps. In the autoregressive phase, we remove the clean frame at the beginning of the window, and append a new noisy frame at the end. This is followed by $k$ denoising steps, yeilding a new clean frame, hence each new frame needs only $k$ forward passes of the model and the autoregressive phase needs $(F-25)k$ forward passes. Finally, in the termination phase all the frames currently in the window get fully denoised. The last frame already being denoised $k$ steps, it takes $24k$ forward passes to finish the termination phase. Summing these phases, our method requires $\frac{F+24}{25}d$ forward passes of the model to generate a $F$ frame video with each frame denoised through $d$ steps. On a GH200 GPU, each forward pass takes around 1 second, initializing the sampler around 20 seconds, and decoding the denoised latent features takes $0.25$ seconds per frame. One could use the above explanation and estimates to calculate the inference time based on their needs. \cref{tab:inference_time} provides specific examples, illustrating the time required to generate videos with 25, 50, and 150 frames, each frame undergoing $d=50$ denoising steps.
\begin{table*}[h!]
\centering
\begin{tabular}{c c c c c}
\toprule
\text{Frames $F$} & \text{Init time (s)} & \text{Sampling time (s)} & \text{Decoding time (s)} & Total time (s) \\
\midrule
25 & 20 & 98 & 6 & 124\\
50 & 20 & 148 & 12 & 180 \\
150 & 20 & 348 & 36 & 404 \\
\bottomrule
\end{tabular}
\caption{Inference time calculation examples for different number of frames.}
\label{tab:inference_time}
\end{table*}

\subsection{Data Curation}
Since our focus is on learning a world model, we emphasize curating data that ensures in-distribution samples with reliable control rather than prioritizing aesthetically pleasing generations. To achieve this, we carefully select filtering methods and thresholds to balance efficiency, quality, and adherence to the desired data distribution.

However, even after filtering based on the aesthetic score, several undesirable samples remain, including overly blurry videos, night recordings with minimal visibility, or clips affected by dirty camera lenses. To address these issues, we additionally utilize PIQE as a distortion detector~\cite{piqe}. While a PIQE score above 50 typically indicates poor quality, the diversity of our dataset—including scenes such as urban environments, rural highways, and night recordings—necessitates a higher threshold to minimize false positives. We therefore set the threshold to 70–80, achieving a balance that minimizes false positives (e.g., retaining valid night driving scenes) while removing the problematic clips mentioned earlier.

\begin{figure*}[h!]
    \centering
    \begin{subfigure}[t]{\textwidth}
        \centering
        \includegraphics[width=\textwidth]{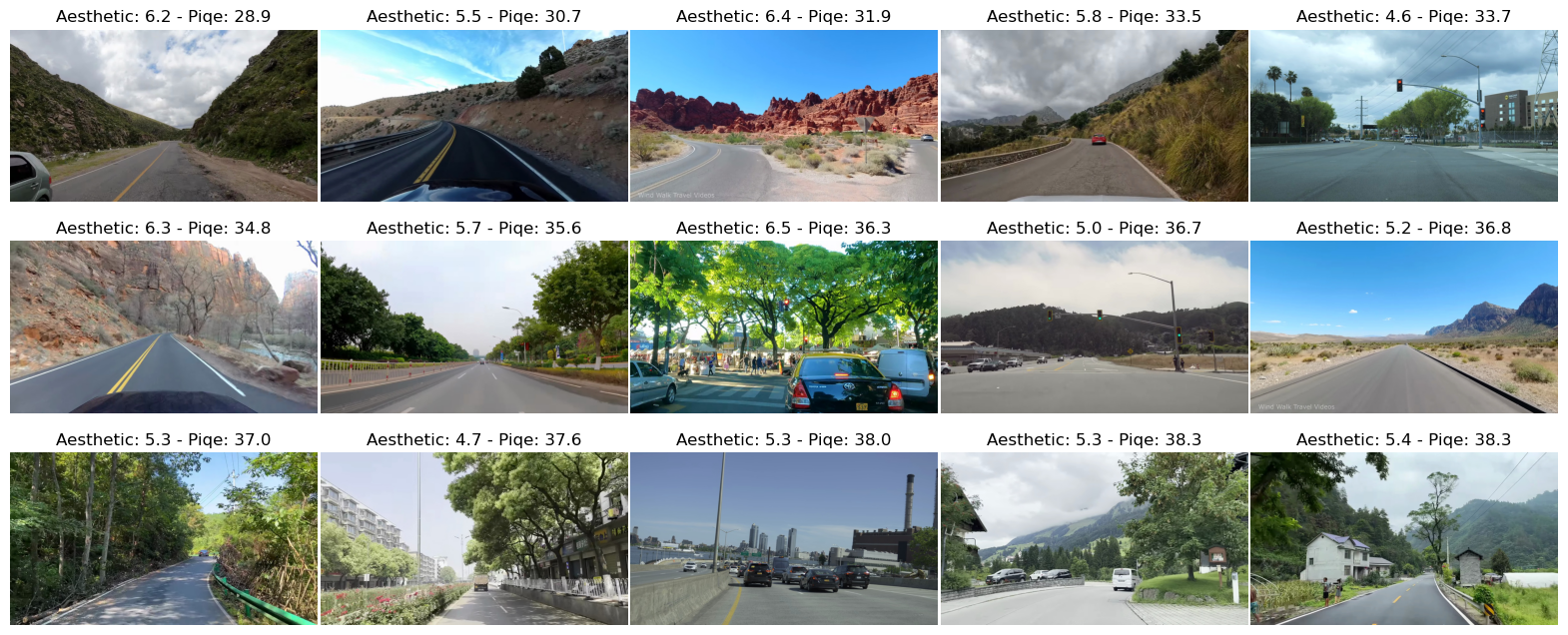}
        \caption{High-quality images, with high aesthetic score ($\geq 4$) and low Piqe score ($\leq 50$).}
        \label{fig:high_quality}
    \end{subfigure}
    \begin{subfigure}[t]{\textwidth}
        \centering
        \includegraphics[width=\textwidth]{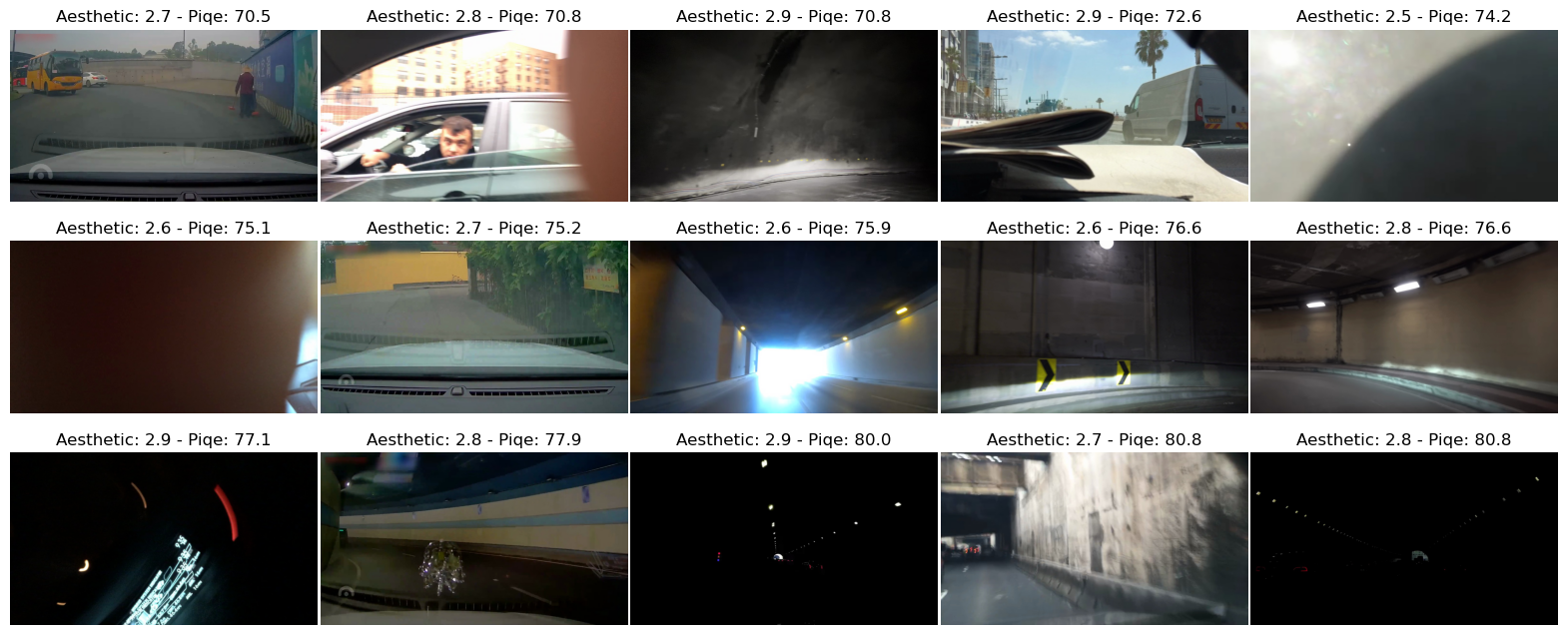}
        \caption{Images filtered with aesthetic score $\leq 3$ and Piqe score $\geq 70$.}
        \label{fig:filtered}
    \end{subfigure}
    \begin{subfigure}[t]{\textwidth}
        \centering
        \includegraphics[width=\textwidth]{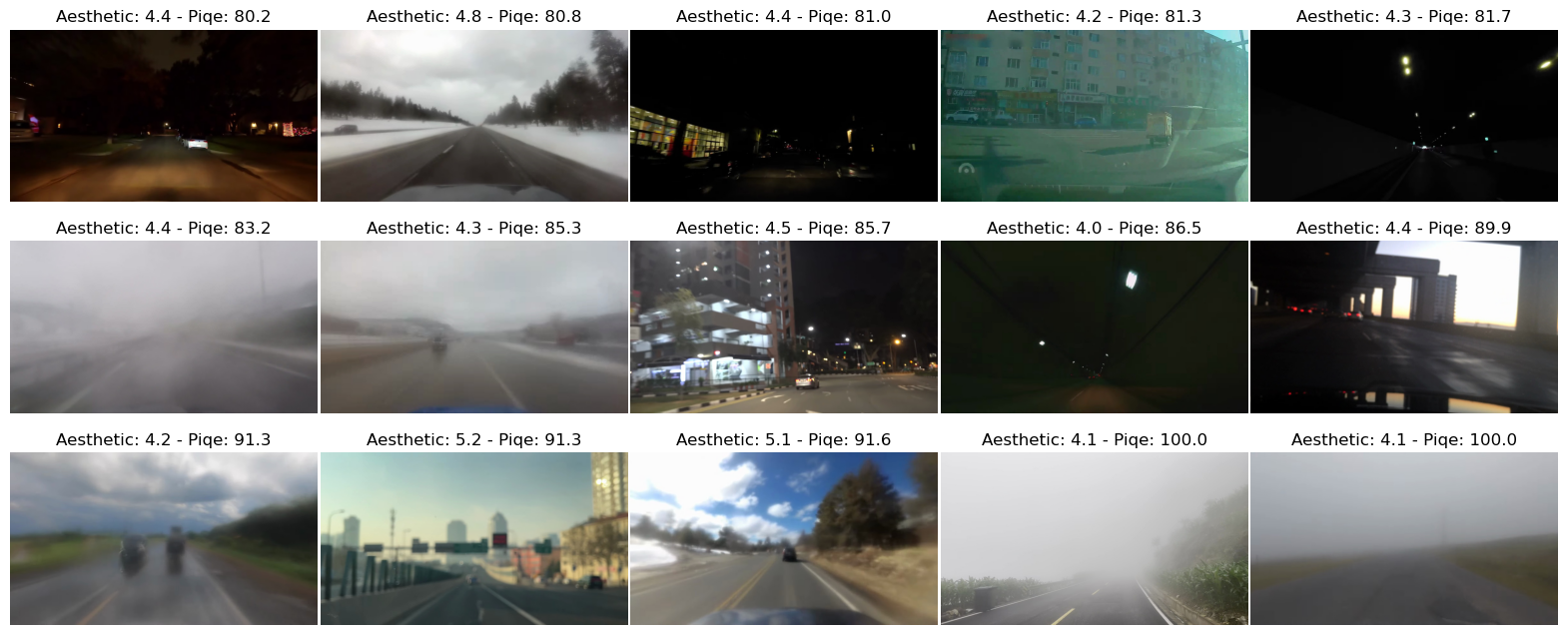}
        \caption{Images with high aesthetic score ($\geq 4$) but high Piqe score ($\geq 80)$.}
        \label{fig:piqe_ae}
    \end{subfigure}
    \caption{Visual examples for quality filtering.}
    \label{fig:quality_filtering}
\end{figure*}

\Cref{fig:quality_filtering} presents both high-quality samples based on the PIQE and aesthetic scores, as well as examples with low-quality scores. Additionally, \Cref{fig:piqe_ae} shows examples of images with a high aesthetic score (indicating good quality) but also a high PIQE score. These results demonstrate that incorporating PIQE into the quality filtering pipeline effectively removes additional unwanted samples.

For both levels of diversity filtering, we employ DINOv2 (large), which we found to outperform alternatives such as CLIP and SSCD~\cite{Pizzi2022ASD} in representing diversity within and across video clips.

\begin{figure*}[h!]
    \centering
    \begin{subfigure}[t]{\textwidth}
        \centering
        \includegraphics[width=\textwidth]{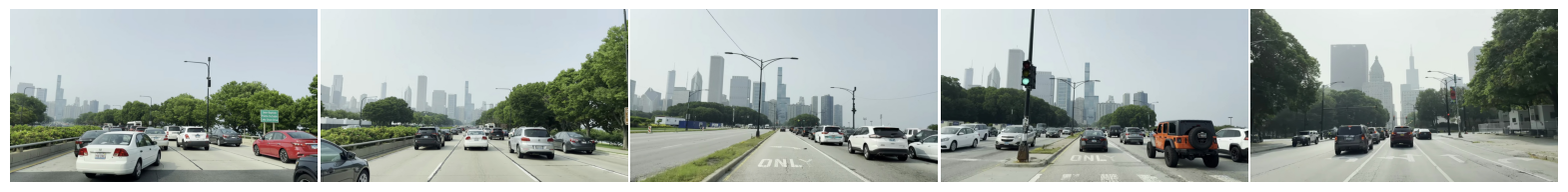}
        \caption{ Images with a cross similarity $\geq0.90$. }
        \label{fig:cross_diversity}
    \end{subfigure}
    \begin{subfigure}[t]{\textwidth}
        \centering
        \includegraphics[width=\textwidth]{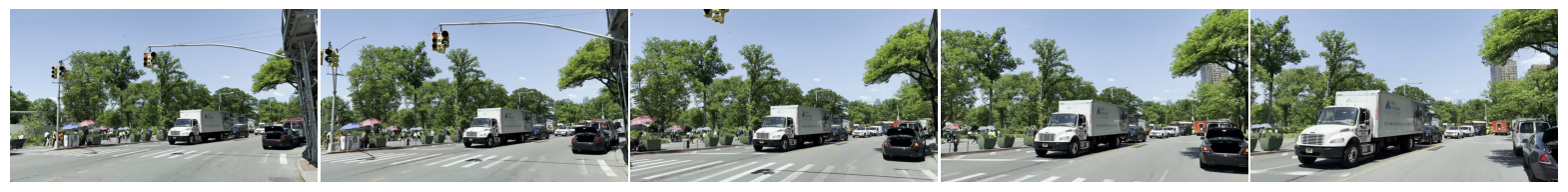}
        \caption{Video clip with high intra-diversity of $ 0.24$.}
        \label{fig:intra_diversity}
    \end{subfigure}
    \begin{subfigure}[t]{\textwidth}
        \centering
        \includegraphics[width=\textwidth]{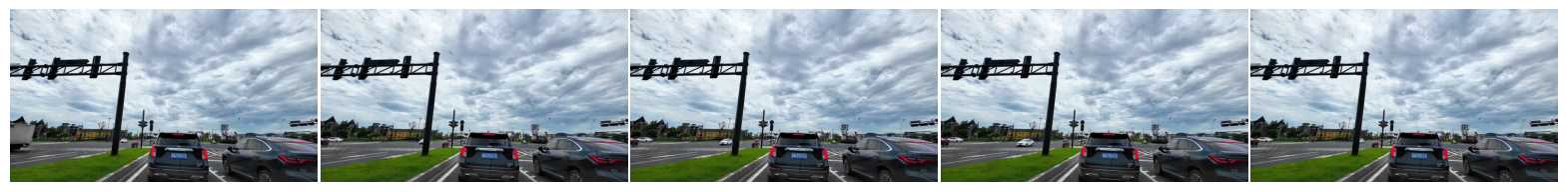}
        \caption{Video clip with low intra-diversity of $\leq 0.02$.}
        \label{fig:low_intra}
    \end{subfigure}
    \begin{subfigure}[t]{\textwidth}
        \centering
        \includegraphics[width=\textwidth]{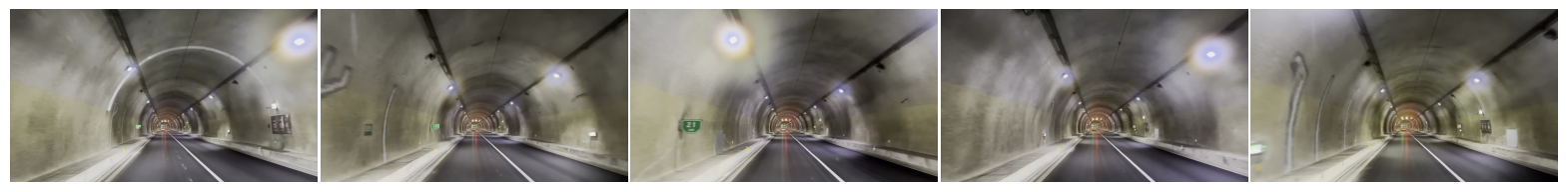}
        \caption{Video clip with low intra-diversity of $\leq 0.02$i, but high motion score (0.12).}
        \label{fig:motion_score}
    \end{subfigure}
    \caption{Visual examples for diversity filtering.}
    \label{fig:diversity_filtering}
\end{figure*}

For cross-clip diversity filtering, we compute the DINO feature vector of the middle frame of each video and calculate the cosine similarity between all resulting vectors. On our dataset, even high thresholds of $0.80$ filtered out entire videos with monotone highway drives featuring little diversity. Consequently, we opted for thresholds between $0.90$ and $0.98$ for our training. Example frames with cross-similarity $\geq 0.9$ are shown in \Cref{fig:cross_diversity}.

For intra-clip diversity, we aim to measure meaningful changes within a clip. In driving videos, the typically high ego-motion makes a motion score based solely on optical flow unsuitable (see examples in \Cref{fig:motion_score}). 

To address this, we process the start and end frames through DINO, extract the feature maps, and compute the cosine similarity between the feature vectors of these frames. We then count the number of tokens with cosine similarity $\leq 0.5$ and normalize by the total number of spatial features. This results in small thresholds (ranging from $0$ to $0.05$) that effectively capture intra-clip diversity for training.

Finally, we observed that DINO occasionally failed to compute meaningful features for certain samples, allowing some static videos to evade filtering. To mitigate this, we additionally apply a motion score based on the average optical flow magnitude between the start and end frames, using a low threshold of $0.02$ to further filter such cases.

\begin{table}[h!]
\centering
\begin{tabular}{c l c c}
\toprule
\text{Stage} & \text{Filter Type} & \text{Threshold} & \text{Data (\%)} \\
\midrule
\multirow{4}{*}{Stage 1} & Aesthetic Score & $\leq 4.0$ & 91\% \\
 & PIQE & $\geq 70$ & 89\% \\
 & Intra- Similarity & $\leq 0.02$ & 80\% \\
 & Motion Score & 0.02 & 79\% \\
 & Cross- Similarity & $\geq 0.98$ & 76\% \\
\hline
\multirow{4}{*}{Stage 2} & Aesthetic Score & 4.2 & 91\% \\
 & PIQE & 70 & 89\% \\
 & Intra- Similarity & 0.02 & 80\% \\
 & Motion Score & 0.02 & 79\% \\
 & Cross- Similarity & 0.95 & 68\% \\
\hline
\bottomrule
\end{tabular}
\caption{Percentage of remaining data after each filter step, starting from 100\%.}
\label{tab:curation}
\end{table}

\section{Implementation Details}

%
As a baseline, we employ H100 GPUs with 100 GB of memory. Due to the increased size of our network, we incorporate activation checkpointing and optimizer sharding to mitigate memory constraints, utilizing the DeepSpeed library \cite{Rasley2020DeepSpeedSO}.

\subsection{Training Stages}
Our training process builds upon the SVD video model \cite{svd} and the EDM framework \cite{Karras2022ElucidatingTD}. To achieve fine-grained, high-quality control, we employ a two-stage training regime, detailed as follows:

\subsubsection{Control Learning Stage}
In this stage, diverse control signals and modalities are introduced. External modules that inject new information into the network are initialized to zero. Given the wide variety of information and tasks across spatial and temporal layers, the entire network is trained without freezing layers or using custom learning rates.

For DINO control, 0 to 10 frames are randomly sampled, a region within these frames is selected, and the regions are encoded using DINOv2 \cite{dinov2}. Following \cite{cage}, tokens are randomly masked to produce 0 to $n_\text{tokens}$ per frame, with $n_\text{tokens}$ set to 16 to maintain sparsity. For identity training, the same frames are used, with 0 to 4 source frames randomly selected and 0 to 3 target frames sampled per source frame, as described in \cref{sec:object_level_control}. Optical flow for the identity training is obtained using RAFT \cite{Teed2020RAFTRA}. 

The initial resolution is \(320 \times 576\), with a learning rate of \(8 \times 10^{-5}\) and an effective batch size of 1024. Training spans two epochs (15k steps), with control learning verified as detailed in \cref{sec:comparison_controllability}. Weak filtering thresholds are applied to maximize training throughput while emphasizing intra-clip diversity to enhance variability in control signals.

\subsubsection{High-Resolution Fine-Tuning}
This stage aims to refine the quality of the control. Training is conducted at a higher resolution of \(576 \times 1024\). As DINO control operates at a downsampling factor of 16, this resolution allows for four times more opportunities for token placement. To maintain sparsity, the number of retained tokens after masking is increased to \(n_\text{tokens} = 32\).

Training continues with a reduced learning rate of \(4 \times 10^{-5}\) and an effective batch size of 512 for one epoch (6k steps). Stricter filtering thresholds are applied during this stage to ensure higher-quality outputs. The thresholds and corresponding data retention percentages for the different training stages are summarized in \cref{tab:curation}.

%
%

\section{Additional Evaluation}
\textbf{Depth Generation Quality.}
\cref{tab:controllability_comparison_obj_out} presents GEM’s depth evaluation using AbsRel and $\delta$, compared to DepthAnything V2’s small and large models. Interestingly, while the training labels are from the small model, results indicate GEM’s depth generations align more closely with the large model on OpenDV, more accurate model in the OpenDV dataset, demonstrating improved depth accuracy over the input.



\begin{table}[h]
    \centering
    \resizebox{\columnwidth}{!}{%
    \begin{tabular}{ccccc}
        \toprule
        & \multicolumn{2}{c}{Nuscenes} & \multicolumn{2}{c}{OpenDV} \\
        & AbsRel  $\downarrow$ & $\delta$  $\uparrow$ &  $AbsRel \downarrow$ &  $\delta$  $\uparrow$ \\ \midrule
        GEM (vs ViT-S) & \textbf{0.17}& \textbf{0.79}& 0.17& 0.8\\
        GEM (vs ViT-L) & 0.2& 0.75& \textbf{0.13}& \textbf{0.84}\\
        \bottomrule
    \end{tabular} }\\
    \caption{Depth generation quality comparison. Our model, despite being trained on pseudo labels from the smaller model, learns to generate more accurate depth maps in the OpenDV dataset, having closer generations to the estimates of the larger DepthAnything model~\cite{depthanything}.}
    \label{tab:controllability_comparison_obj_out}
    
\end{table}

\section{Ablation Studies}
\begin{figure*}[h]
    \centering
    \begin{subfigure}[b]{0.95\textwidth}
        \centering
        \includegraphics[width=\textwidth]{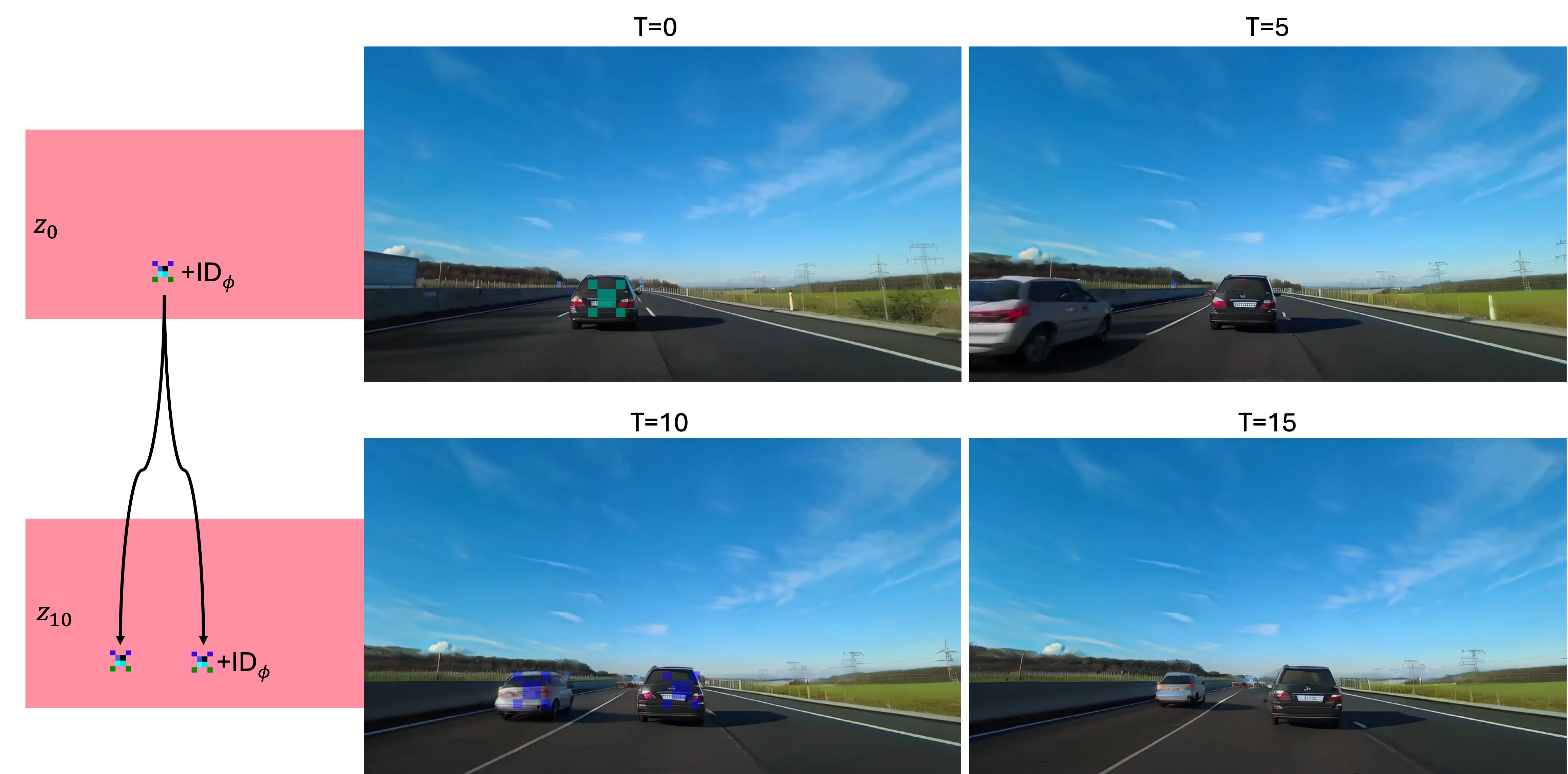}
        \caption{We move the car to the right while inserting another car to the left.}
        \label{fig:sub1}
    \end{subfigure}
    \hfill
    \begin{subfigure}[b]{0.95\textwidth}
        \centering
        \includegraphics[width=\textwidth]{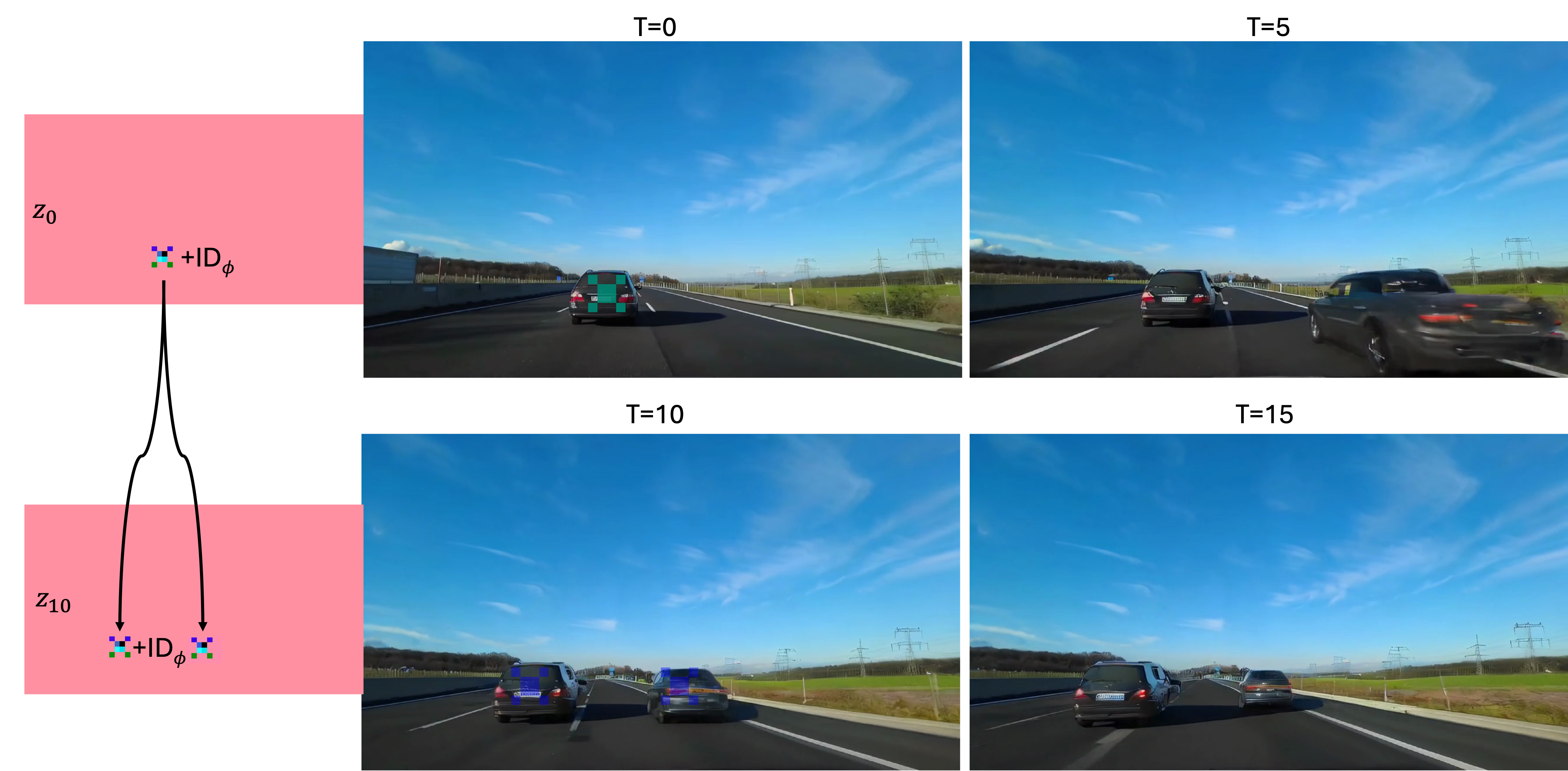}
        \caption{We move the car to the left while inserting another car to the right.}
        \label{fig:sub2}
    \end{subfigure}
    \caption{Demonstration of moving an object while simultaneously inserting a new one nearby. We utilize DINO tokens of the car from the initial frame and replicate them at specified locations and times (e.g., $T=0$ and $T=10$). Identity is added to tokens corresponding across time. The DINO control is shown on the left, and the resulting generation is displayed on the right.}

    \label{fig:examples_id}
\end{figure*}

\textbf{Identity Evaluation}
Showing the significance of adding ID embeddings to the DINO tokens of different objects is challenging. This is because the ID is primarily beneficial in scenarios with ambiguous actions (e.g. two very close objects or when moving an object and inserting another). Therefore, we randomly chose a subset of 100 videos where we can test the importance of adding ID labels. As shown in \cref{tab:id-ablation}, adding ID embeddings resulted in a slight decrease in the Controllability of Object Manipulation metric (COM) error, from 22.4 pixels to 21 pixels. However, COM is not an ideal metric for evaluating the role of ID embeddings in these scenarios. To better illustrate their importance, we provide examples in \cref{fig:examples_id}. These highlight the critical role of ID embeddings in resolving ambiguities and enabling more precise control when managing interactions with adding different controls on different objects. 
\begin{table}[h]
    \centering

    \begin{tabular}{cc}
        \toprule
        & \multicolumn{1}{c}{OpenDV} \\
        & COM $\downarrow$ \\ \midrule
        With object ID & 21.0  \\
        Without object ID  & 22.4 \\
        \bottomrule
    \end{tabular} \\
    \caption{Comparison of adding ID embeddings for DINO tokens.}
    \label{tab:id-ablation}
    
\end{table}

\section{Qualitative Results}
\begin{figure}[htbp]
    \centering
    \begin{subfigure}{0.95\textwidth}
        \centering
        \includegraphics[width=\linewidth]{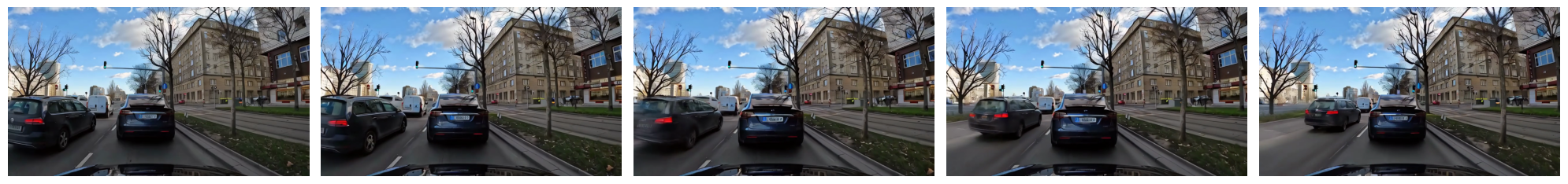}
        \label{fig:sub3}
    \end{subfigure}
    \vspace{1em} 

    \begin{subfigure}{0.95\textwidth}
        \centering
        \includegraphics[width=\linewidth]{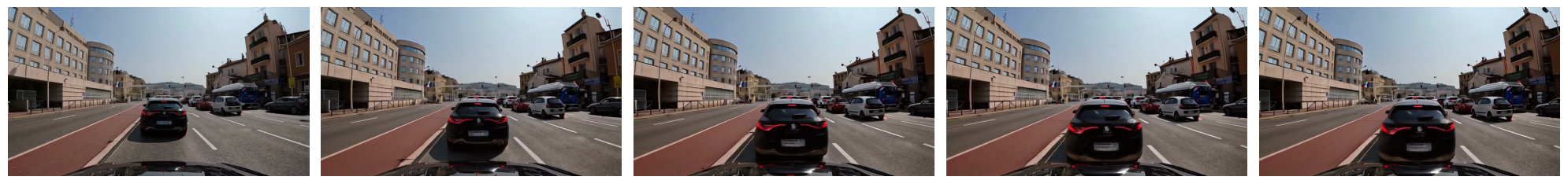}
        \label{fig:sub4}
    \end{subfigure}
    
    \vspace{1em} 

    \begin{subfigure}{0.95\textwidth}
        \centering
        \includegraphics[width=\linewidth]{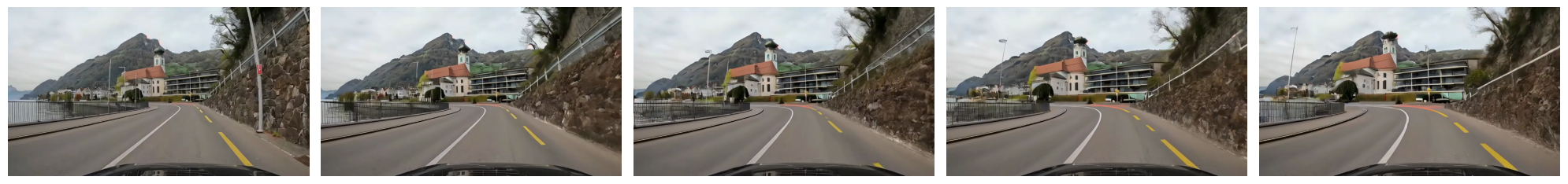}
        \label{fig:sub5}
    \end{subfigure}
    
    \vspace{1em}
    \begin{subfigure}{0.95\textwidth}
        \centering
        \includegraphics[width=\linewidth]{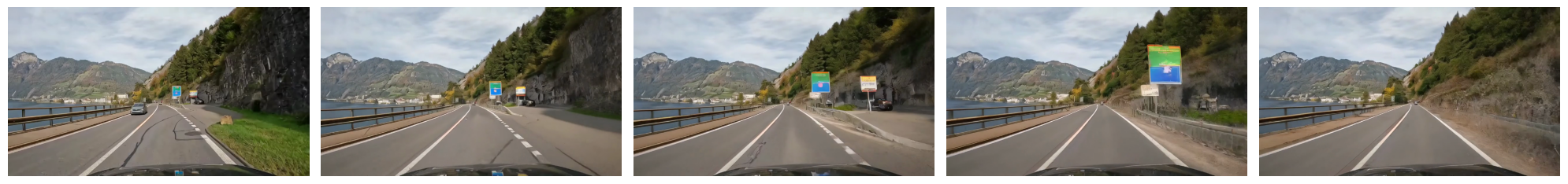}
        \label{fig:sub6}
    \end{subfigure}

    \vspace{1em}

    \begin{subfigure}{0.95\textwidth}
        \centering
        \includegraphics[width=\linewidth]{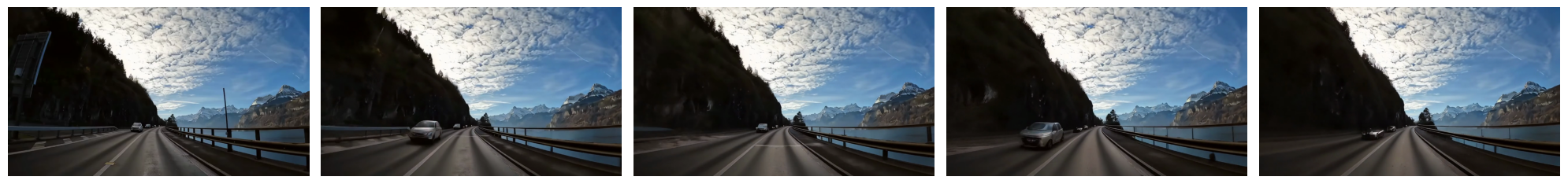}
        \label{fig:sub7}
    \end{subfigure}
    
    \vspace{1em}
    \begin{subfigure}{0.95\textwidth}
        \centering
        \includegraphics[width=\linewidth]{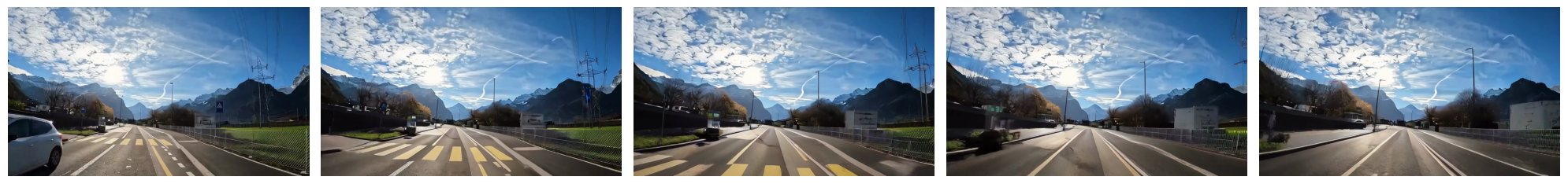}
        \label{fig:sub8}
    \end{subfigure}
    
    \vspace{1em} 

    \begin{subfigure}{0.95\textwidth}
        \centering
        \includegraphics[width=\linewidth]{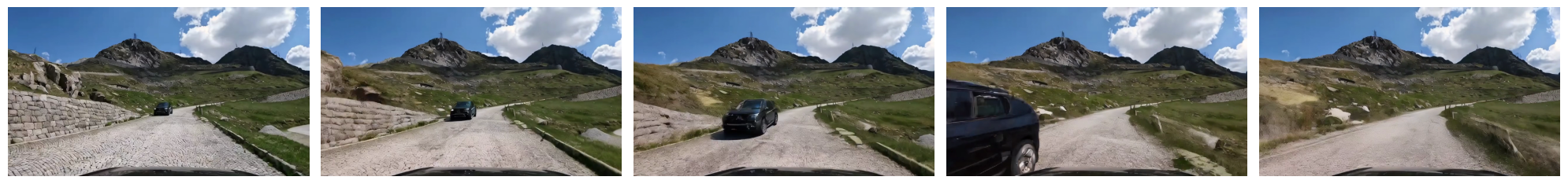}
        \label{fig:sub9}
    \end{subfigure}

    \vspace{1em}

    \begin{subfigure}{0.95\textwidth}
        \centering
        \includegraphics[width=\linewidth]{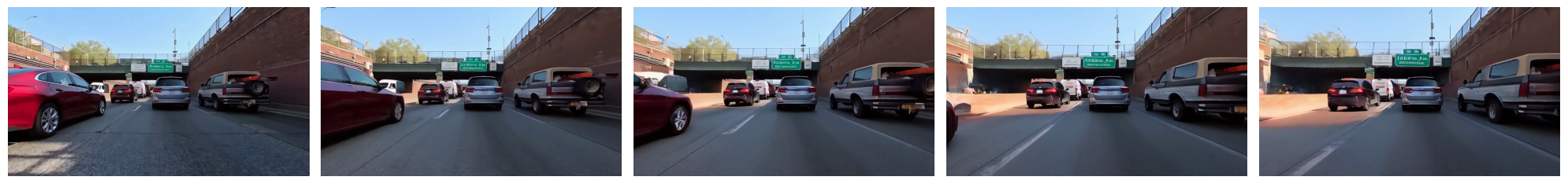}
    \end{subfigure}

    \caption{Generated videos with 25 frames on OpenDV.}
    \label{fig:generated_frames}
\end{figure}

\begin{figure*}[htbp]
    \centering
    \begin{subfigure}{0.95\textwidth}
        \centering
        \includegraphics[width=\linewidth]{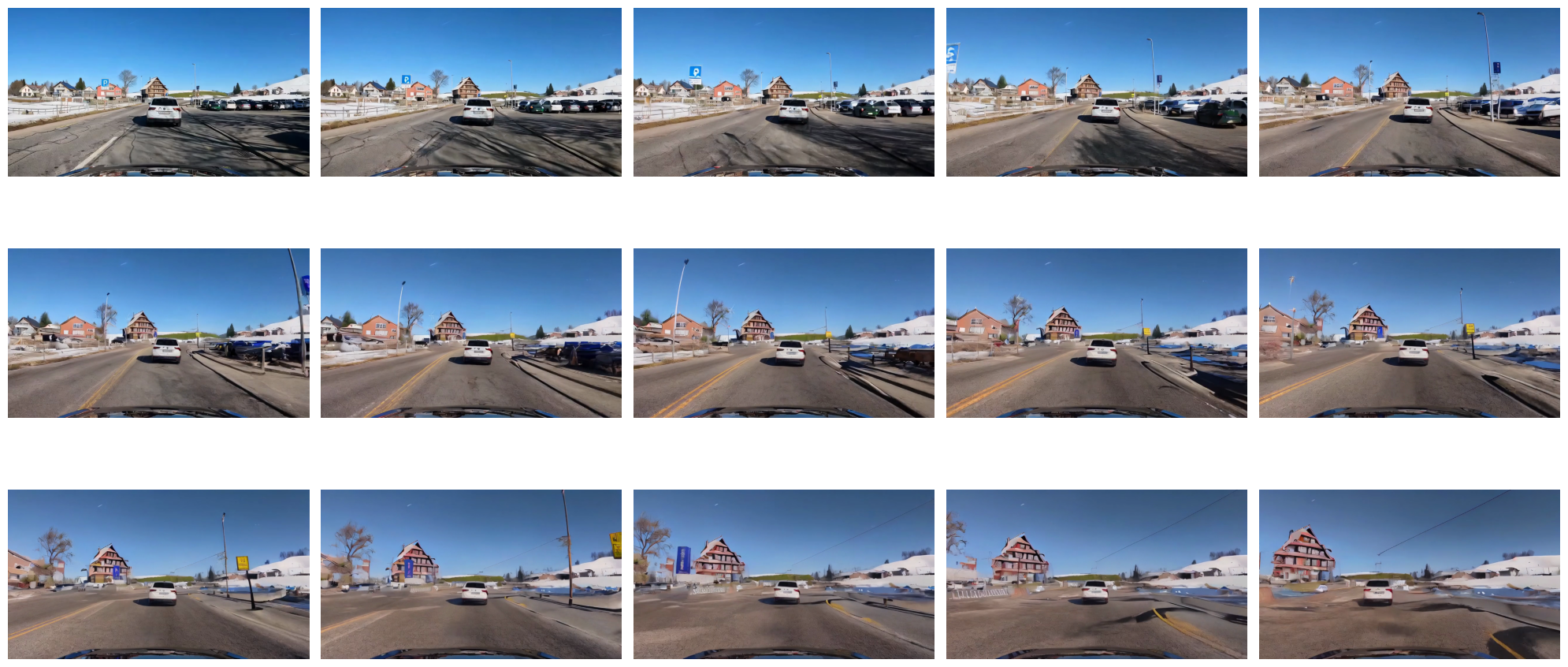}
        \label{fig:first}
    \end{subfigure}
    \hfill
    \begin{subfigure}{0.95\textwidth}
        \centering
        \includegraphics[width=\linewidth]{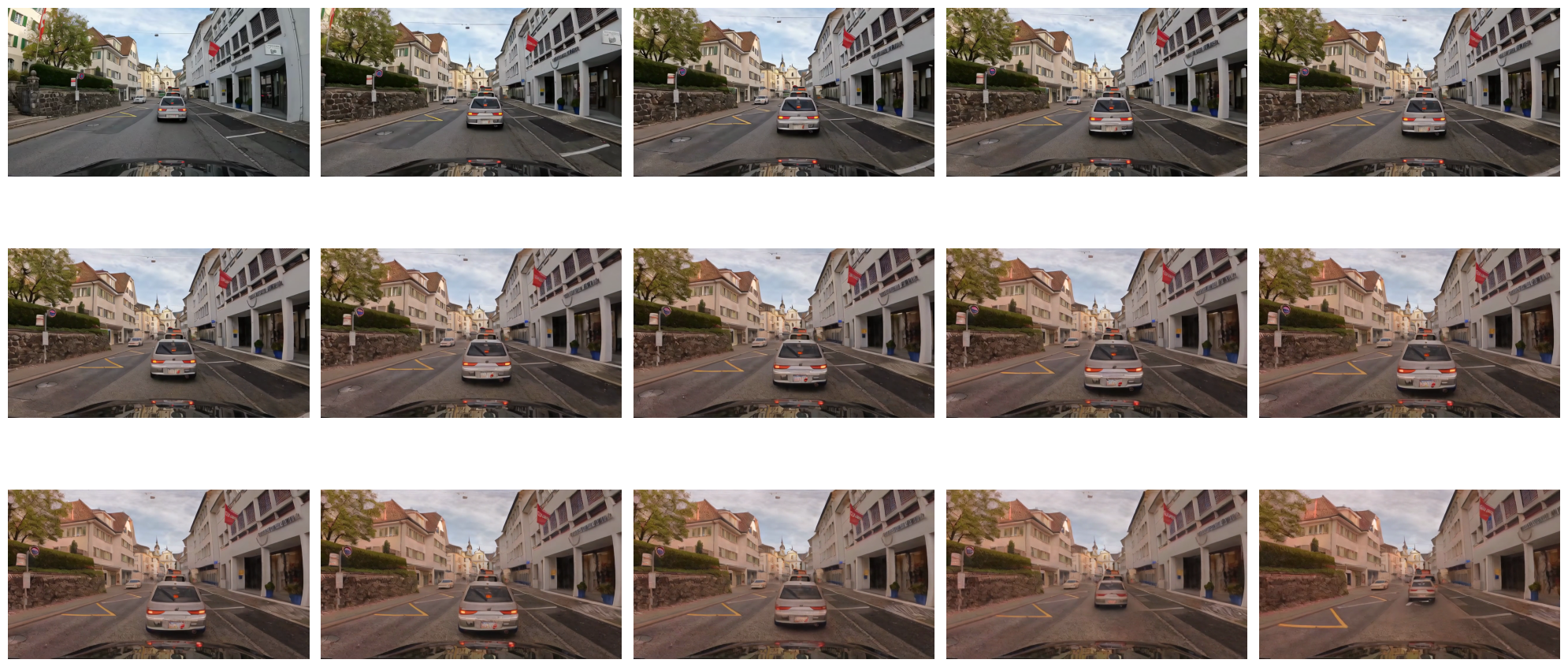}
        \label{fig:second}
    \end{subfigure}
    \caption{Generated videos with 150 frames on OpenDV.}
    \label{fig:combined_long_gen}
\end{figure*}

\begin{figure*}[t]
    \centering
    \begin{subfigure}{0.95\textwidth}
        \centering
        \includegraphics[width=\linewidth]{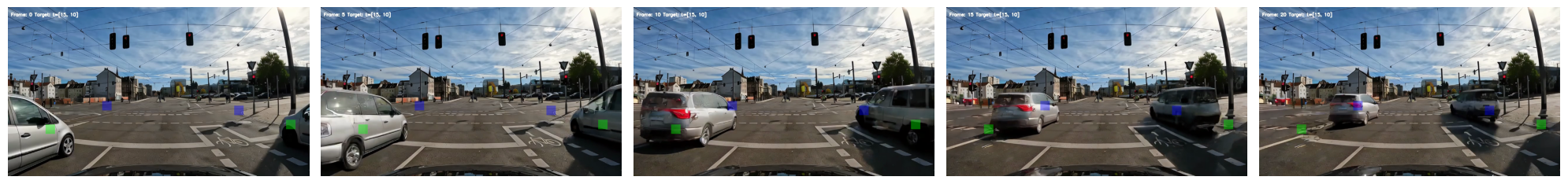}
        \caption{Moving two cars.}
        \label{fig:sub1}
    \end{subfigure}
    
    \vspace{1em} 
    
    %
    
    %
    
    \begin{subfigure}{0.95\textwidth}
        \centering
        \includegraphics[width=\linewidth]{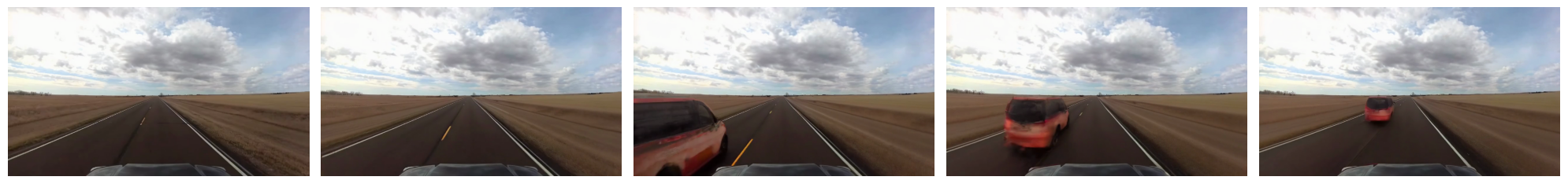}
        \caption{Inserting a car on the left.}
        \label{fig:sub4}
    \end{subfigure}
    
    \vspace{1em}
    
    
    %
    \begin{subfigure}{0.95\textwidth}
        \centering
        \includegraphics[width=\linewidth]{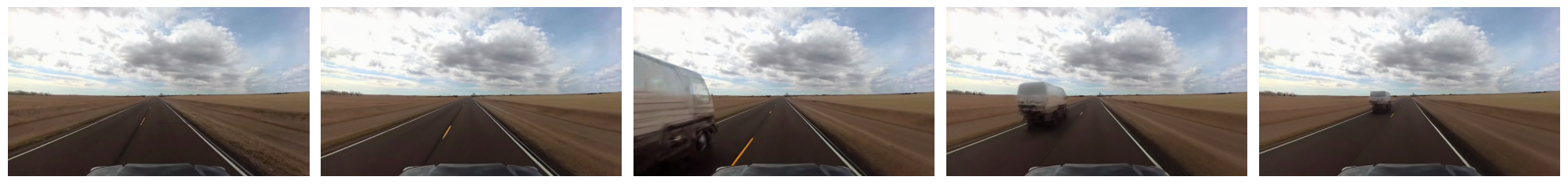}
        \caption{Inserting a truck on the left.}
        \label{fig:sub6}
    \end{subfigure}
    %
    %
    %
    %
    
    \caption{Examples of moving and inserting objects with DINO control.}
    \label{fig:control_frames}
\end{figure*}

\begin{figure*}[t]
    \centering
    \begin{subfigure}{0.95\textwidth}
        \centering
        \includegraphics[width=\linewidth]{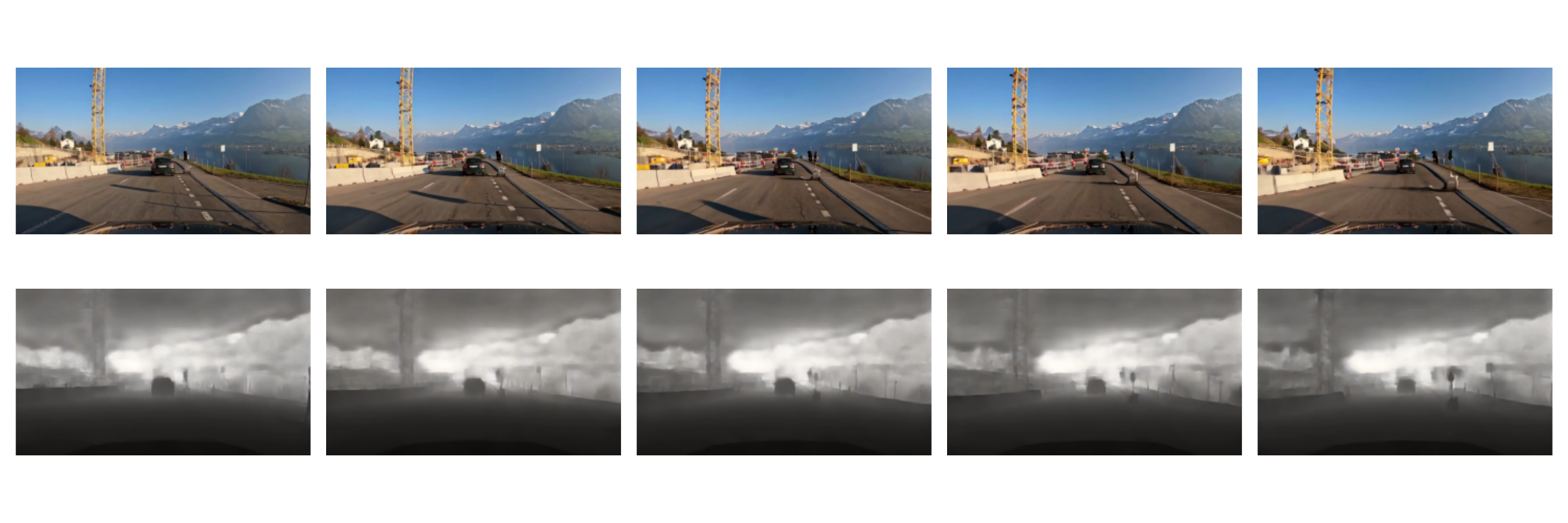}
        \label{fig:first}
    \end{subfigure}
    \hfill
    \begin{subfigure}{0.95\textwidth}
        \centering
        \includegraphics[width=\linewidth]{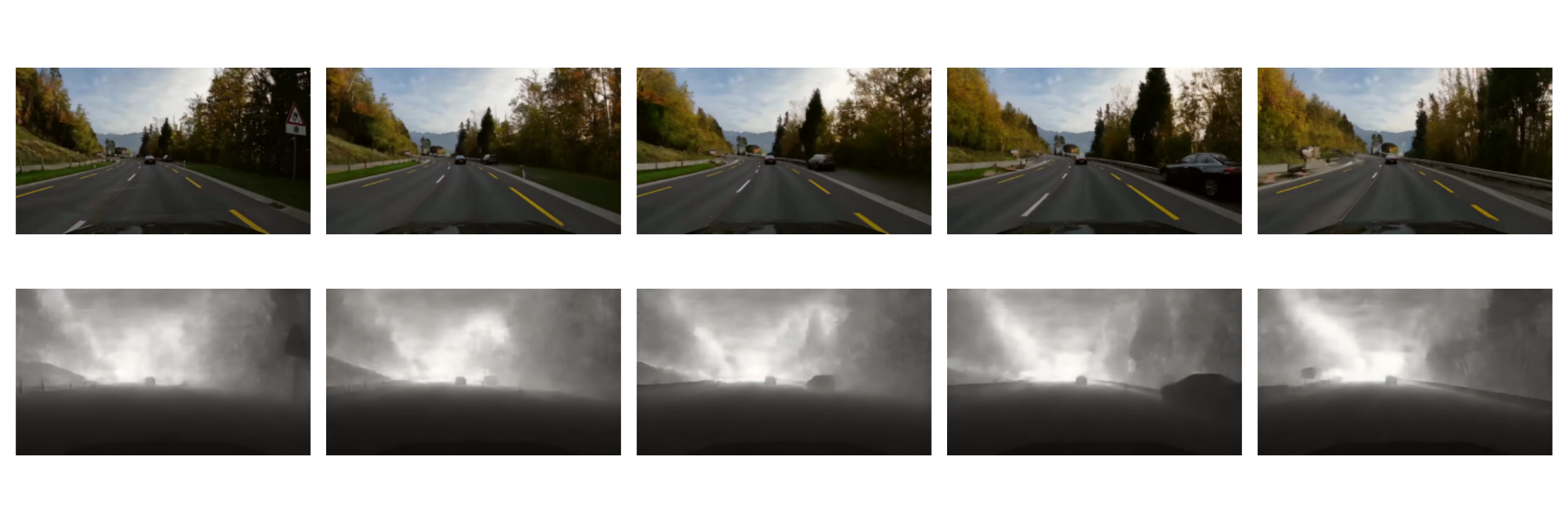}
        \label{fig:second}
    \end{subfigure}
    \caption{Multimodal generations.}
    \label{fig:combined_multimodal_vis}
\end{figure*}

\cref{fig:generated_frames,fig:control_frames,fig:combined_long_gen,fig:combined_multimodal_vis} show qualitative examples of our generations, our controls, long generation and multimodal outputs. 


\end{document}